\documentclass[preprint,10pt]{elsarticle}

\usepackage{amsmath,amssymb,amsfonts}
\usepackage{mathtools}
\usepackage{bm}
\usepackage{graphicx}
\usepackage{booktabs}
\usepackage{enumitem}
\usepackage{hyperref}
\usepackage{cleveref}
\usepackage{xcolor}
\usepackage{lineno}
\usepackage{subcaption}
\usepackage{todonotes}

\modulolinenumbers[5]

\journal{Journal of Computational Physics}
\begin{document}
\begin{frontmatter}
\title{
Uncertainty Quantification in PINNs for Turbulent Flows: Bayesian Inference and Repulsive Ensembles}
\author[Brown]{Khemraj Shukla}
\author[CalTech]{Zongren Zou}
\author[MIT]{Theo Kaeufer}
\author[MIT]{Michael Triantafyllou}
\author[Brown]{George Em Karniadakis\corref{cor}}

\cortext[cor]{Corresponding author}

\affiliation[Brown]{organization={Division of Applied Mathematics, Brown University},
            city={Providence},
            state={RI},
            country={USA}}

\affiliation[CalTech]{organization={Department of Computing and Mathematical Sciences, California Institute of Technology},
            city={Pasadena},
            state={CA},
            country={USA}}

\affiliation[MIT]{organization={Department of Mechanical Engineering, Massachusetts Institute of Technology},
            city={Cambridge},
            state={MA},
            country={USA}}

\begin{abstract}
Physics-informed neural networks (PINNs) have emerged as a promising framework for solving inverse problems governed by partial differential equations (PDEs), including the reconstruction of turbulent flow fields from sparse data. However, most existing PINN formulations are deterministic and do not provide reliable quantification of epistemic uncertainty, which is critical for ill-posed problems such as data-driven Reynolds-averaged Navier–Stokes (RANS) modeling.
In this work, we develop and systematically evaluate a set of probabilistic extensions of PINNs for uncertainty quantification in turbulence modeling. The proposed framework combines (i) Bayesian PINNs with Hamiltonian Monte Carlo sampling and a tempered multi-component likelihood, (ii) Monte Carlo dropout, and (iii) repulsive deep ensembles that enforce diversity in function space. Particular emphasis is placed on the role of ensemble diversity and likelihood tempering in improving uncertainty calibration for PDE-constrained inverse problems.
The methods are assessed on a hierarchy of test cases, including the Van der Pol oscillator and turbulent flow past a circular cylinder at Reynolds numbers Re = 3,900 (direct numerical simulation data) and Re = 10,000 (experimental particle image velocimetry data). The results demonstrate that Bayesian PINNs provide the most consistent uncertainty estimates across all inferred quantities, while function-space repulsive ensembles offer a computationally efficient approximation with competitive accuracy for primary flow variables. These findings provide quantitative insight into the trade-offs between accuracy, computational cost, and uncertainty calibration in physics-informed learning, and offer practical guidance for uncertainty quantification in data-driven turbulence modeling.

\end{abstract}

\begin{keyword}
Physics-informed neural networks \sep Uncertainty quantification \sep Bayesian inference \sep Repulsive deep ensembles  \sep Reynolds-averaged Navier-Stokes equations
\end{keyword}

\end{frontmatter}

% ====================================================================
\section{Introduction}
\label{sec:introduction}

Among the hierarchy of turbulence modeling approaches, the Reynolds-Averaged Navier--Stokes (RANS) equations remain the most widely adopted framework in industrial applications due to their favorable balance between computational efficiency and predictive capability~\cite{wilcox2006turbulence,pope2000turbulent}. By solving for time-averaged flow quantities, RANS models avoid resolving the full spectrum of turbulent scales; however, this simplification introduces a fundamental closure problem associated with the modeling of the Reynolds stress tensor. In practice, this term is approximated through empirical or semi-empirical turbulence models, which can lead to significant inaccuracies for flows involving strong separation, streamline curvature, or non-equilibrium effects~\cite{durbin2018some,spalart2009detached}.

Although higher-fidelity approaches such as Large Eddy Simulation (LES) and Direct Numerical Simulation (DNS) have become increasingly feasible with modern computational resources, their application remains limited to canonical configurations or moderate Reynolds numbers due to their substantial computational cost~\cite{moin1998direct,sagaut2006large}. In many engineering contexts---particularly in design optimization and uncertainty quantification---fast turnaround times and the need for repeated evaluations across large parameter spaces render RANS-based methods indispensable. Consequently, there is a growing need to improve the predictive accuracy of RANS models and to rigorously quantify their associated uncertainties, especially in safety-critical and high-performance applications~\cite{duraisamy2019turbulence,ling2016reynolds}. Addressing these challenges is essential for enhancing the reliability and robustness of CFD as a predictive engineering tool.
 
A central challenge in RANS modeling is that the Reynolds stress closure introduces model-form errors that are difficult to quantify \emph{a priori}. Traditional approaches to turbulence model uncertainty---such as eigenvalue perturbation methods~\cite{emory2013modeling, iaccarino2017eigenspace}---operate within the framework of prescribed closure models and can only explore a limited subspace of possible Reynolds stress realizations. These methods provide useful bounds on predictive uncertainty but cannot systematically incorporate observational data to reduce that uncertainty in a principled manner.
 
In recent years, Physics-Informed Neural Networks (PINNs)~\cite{raissi2019physics, shukla2020physics, shukla2021parallel} have emerged as a promising data-driven alternative for solving forward and inverse problems governed by partial differential equations. By embedding the governing equations directly into the loss function, PINNs enforce physical consistency while leveraging observational data to reconstruct unknown fields. Several studies have demonstrated the potential of PINNs for mean flow reconstruction in turbulent flows. Patel et al.~\cite{patel2024turbulence} showed that PINNs can infer velocity and pressure fields from sparse measurements by treating the Reynolds stress divergence as an unknown source term in the RANS equations. Eivazi et al.~\cite{eivazi2022physics} extended this idea to reconstruct turbulent channel flows, and Zhang et al.~\cite{zhang2025turbulence} explored PINN-based closure modeling with experimental data. These works convincingly demonstrate that PINNs can recover accurate mean flow fields from limited observations.
 
However, existing PINN-based turbulence modeling approaches share a critical limitation: they treat the neural network as a deterministic surrogate and provide no systematic quantification of the epistemic uncertainty inherent in the inverse problem. This omission is particularly consequential in the turbulence modeling context, where the inverse problem is fundamentally ill-posed---multiple combinations of velocity, pressure, and Reynolds stress fields can satisfy both the governing equations and the sparse observations to comparable accuracy. Without uncertainty estimates, it is impossible to distinguish regions where the PINN predictions are well-constrained by data from regions where they are driven primarily by the implicit inductive bias of the network architecture and training procedure \cite{psaros2023uncertainty, yang2021b, zou2025uncertainty, liu2024flow, meng2022learning, zou2025multi}. Furthermore, deterministic PINNs offer no guidance on the marginal value of additional measurements, limiting their utility for experimental design and adaptive data collection \cite{meng2021multi, aikawa2024improving, zou2024leveraging}.
 
The broader scientific-machine learning literature offers several approaches for uncertainty quantification in neural networks, each with distinct trade-offs \cite{psaros2023uncertainty, linka2022bayesian} for a detailed discussion. Bayesian neural networks~\cite{neal1996bayesian} provide a principled probabilistic framework by placing prior distributions over the network weights and computing posterior distributions conditioned on the data. Hamiltonian Monte Carlo (HMC)~\cite{neal2011mcmc} and its adaptive variant, the No-U-Turn Sampler (NUTS)~\cite{hoffman2014no}, enable asymptotically exact posterior sampling but are computationally expensive in high-dimensional parameter spaces. Monte Carlo  Dropout~\cite{gal2016dropout} offers a computationally efficient alternative by interpreting dropout at test time as approximate variational inference, though the resulting uncertainty estimates are limited by the expressiveness of the induced variational family. Deep ensembles~\cite{lakshminarayanan2017simple} provide well-calibrated predictions through the diversity of independently trained networks, but standard ensembles trained from random initializations often converge to similar solutions, leading to underestimation of epistemic uncertainty.
 
To address the challenge of ensemble collapse, D'Angelo and Fortuin~\cite{d2021repulsive} introduced Repulsive Deep Ensembles, which incorporate kernel-based repulsive interactions between ensemble members to enforce diversity during training. They showed that repulsive ensembles can be interpreted as approximating Bayesian posterior inference through Wasserstein gradient flow dynamics. Pilar et al.~\cite{pilar2025repulsive} further extended this framework, demonstrating its effectiveness in scientific computing applications. However, repulsive ensembles have not yet been applied to PINNs for systems of PDEs, nor have they been systematically compared with Bayesian sampling methods in the context of turbulence modeling.

In this work, we present a comprehensive framework for uncertainty quantification in PINN-based turbulence modeling that addresses the limitations identified above. We develop, implement, and systematically compare three complementary methods---Bayesian PINNs (BPINNs), MC Dropout PINNs, and Repulsive Deep Ensemble PINNs---applied to the inverse RANS problem of jointly inferring velocity, pressure, and Reynolds stress fields from sparse observations.

\paragraph{Main contributions}  
The key contributions of this work are summarized as follows:

\begin{enumerate}[leftmargin=*]

\item We propose a repulsive deep ensemble framework for physics-informed neural networks  tailored to systems of PDEs. This approach enables scalable uncertainty quantification for Reynolds stress and mean flow inference under sparse and localized observations, while directly addressing the inherent ill-posedness of PDE-constrained inverse problems in turbulence modeling.

\item We introduce and systematically study repulsion mechanisms in both function space and parameter space. Function-space repulsion promotes diversity in predicted flow fields and Reynolds stresses by penalizing similarity in network outputs, whereas parameter-space repulsion drives exploration across distinct regions of the weight landscape. Their combination mitigates ensemble collapse and enhances posterior exploration relative to standard deep ensembles.

\item We demonstrate the effectiveness of the proposed framework in a partial observation setting, where measurements are confined to a localized region near a cylinder. The method successfully recovers multiple physically consistent solutions that satisfy both governing equations and sparse noisy data, while the ensemble spread yields meaningful uncertainty estimates in unobserved regions.

\item To overcome the limitations of the original BPINN formulation method \cite{yang2021b}---notably unreliable uncertainty quantification in highly underdetermined settings and poor scalability---we develop a enhanced BPINN with a tempered multi-component likelihood and post-hoc recalibration. Posterior inference is performed using NUTS, with distinct tempering exponents assigned to velocity data, Reynolds stress data, and PDE residuals to reflect their heterogeneous contributions. This improves both sampling efficiency and uncertainty calibration. A post-hoc recalibration step further corrects the global uncertainty scale while preserving spatial structure.

\item We present a comprehensive comparative study across four methods, BPINNs \cite{yang2021b}, MC Dropout \cite{zhang2019quantifying}, and repulsive deep ensembles, on turbulent flow past a cylinder at $\mathrm{Re} = 3{,}900$ (DNS data) and $\mathrm{Re} = 10{,}000$ (experimental data). The comparison highlights trade-offs in computational cost, calibration quality, and fidelity of posterior approximation.

\end{enumerate}
 
The remainder of this paper is organized as follows. Section~\ref{sec:methodology} presents the governing equations and the three uncertainty quantification methods. Section~\ref{sec:experiments} describes the computational experiments, starting with a pedagogical but relevant example of the Van der Pol oscillator benchmark and continuing with the cylinder flow test cases. Section~\ref{sec:conclusions} summarizes the findings and discusses directions for future work.

% ====================================================================
\section{Methodology}
\label{sec:methodology}

% ------------------------------------------------------------------
\subsection{Problem Formulation: Reynolds-Averaged Navier--Stokes Equations}
\label{sec:governing_equations}

We consider the steady-state Reynolds-Averaged Navier--Stokes (RANS) equations for incompressible flow:
\begin{align}
    U \frac{\partial U}{\partial x} + V \frac{\partial U}{\partial y} + \frac{\partial P}{\partial x} - \frac{1}{\mathrm{Re}} \left( \frac{\partial^2 U}{\partial x^2} + \frac{\partial^2 U}{\partial y^2} \right) + f_x &= 0, \label{eq:mom_x} \\
    U \frac{\partial V}{\partial x} + V \frac{\partial V}{\partial y} + \frac{\partial P}{\partial y} - \frac{1}{\mathrm{Re}} \left( \frac{\partial^2 V}{\partial x^2} + \frac{\partial^2 V}{\partial y^2} \right) + f_y &= 0, \label{eq:mom_y} \\
    \frac{\partial U}{\partial x} + \frac{\partial V}{\partial y} &= 0, \label{eq:continuity}
\end{align}
where $U$ and $V$ are the mean velocity components, $P$ is the mean pressure, $\mathrm{Re}$ is the Reynolds number, and $f_x$, $f_y$ represent the divergence of the Reynolds stress tensor and represented as Reynolds forces. Rather than prescribing a turbulence closure model, we treat $f_x$ and $f_y$ as unknown fields to be inferred jointly with the velocity and pressure fields from sparse observational data and the governing equations. We denote the PDE residuals of the $x$-momentum, $y$-momentum, and continuity equations as $r_1$, $r_2$, and $r_3$, respectively.

We parameterize the solution fields using a fully connected feedforward neural network $\mathcal{N}_\theta : \mathbb{R}^2 \to \mathbb{R}^5$ with parameters $\theta$, which maps spatial coordinates $(x, y)$ to the five output quantities $(U, V, P, f_x, f_y)$. All spatial derivatives required for evaluating the PDE residuals are computed via automatic differentiation through the neural network. Input coordinates are normalized to zero mean and unit variance to improve conditioning \cite{raissi2019physics}.

The following subsections describe three approaches for uncertainty quantification in this physics-informed setting: BPINNs (Section~\ref{sec:bpinn}), MC Dropout PINNs (Section~\ref{sec:mc_dropout}), and Repulsive Deep Ensemble PINNs (Section~\ref{sec:rde}).

% ------------------------------------------------------------------
\subsection{Bayesian Physics-Informed Neural Network}
\label{sec:bpinn}

The Bayesian formulation places a prior distribution over the network parameters $\theta$ introduced in Section~\ref{sec:governing_equations} and updates it via Bayes' theorem to obtain a posterior distribution conditioned on both observational data and physics constraints \cite{yang2021b}. The posterior distribution is given by
\begin{equation}
    p(\theta \mid \mathcal{D}) \propto p(\mathcal{D} \mid \theta) \, p(\theta),
    \label{eq:posterior}
\end{equation}
where $p(\theta) = \mathcal{N}(\theta \mid 0, \sigma_\mathrm{prior}^2 \mathbf{I})$ is an isotropic Gaussian prior with $\sigma_\mathrm{prior} = 2.0$, and the likelihood $p(\mathcal{D} \mid \theta)$ encodes both data fidelity and physics constraints.
where $p(\theta)$ denotes prior (in this work, we use the isotropic Gaussian prior, i.e., $\mathcal{N}(\theta \mid 0, \sigma_\mathrm{prior}^2 \mathbf{I})$ with $\sigma_{\text{prior}}=2.0$), and $p(\mathcal{D} \mid \theta)$ denotes the likelihood, in which both data fidelity and physics constraints are encoded.

\subsubsection{Likelihood Construction}
\label{sec:likelihood}

The likelihood comprises three components corresponding to velocity observations, Reynolds stress observations, and physics residuals.

\paragraph{Velocity data likelihood}
Given $N_d$ observation points $\{(x_i, y_i)\}_{i=1}^{N_d}$ with measured velocities $u_i^\mathrm{obs}$ and $v_i^\mathrm{obs}$, we define
\begin{equation}
    \log p_{UV}(\mathcal{D}_{UV} \mid \theta) = -\frac{1}{2} \sum_{i=1}^{N_d} \left[ \left( \frac{U_\theta(x_i, y_i) - U_i^\mathrm{obs}}{\sigma_u} \right)^2 + \left( \frac{V_\theta(x_i, y_i) - V_i^\mathrm{obs}}{\sigma_v} \right)^2 \right],
    \label{eq:lik_uv}
\end{equation}
where $\sigma_u$ and $\sigma_v$ are per-component noise standard deviations.

\paragraph{Reynolds stress data likelihood}
At the same observation points, measurements of the Reynolds stress divergence components $f_{x,i}^\mathrm{obs}$ and $f_{y,i}^\mathrm{obs}$ are incorporated as
\begin{equation}
    \log p_f(\mathcal{D}_f \mid \theta) = -\frac{1}{2} \sum_{i=1}^{N_d} \left[ \left( \frac{f_{x,\theta}(x_i, y_i) - f_{x,i}^\mathrm{obs}}{\sigma_{f_x}} \right)^2 + \left( \frac{f_{y,\theta}(x_i, y_i) - f_{y,i}^\mathrm{obs}}{\sigma_{f_y}} \right)^2 \right].
    \label{eq:lik_f}
\end{equation}

\paragraph{Physics residual likelihood}
At $N_c$ collocation points distributed throughout the computational domain, the RANS residuals are penalized:
\begin{equation}
    \log p_r(\mathcal{R} \mid \theta) = -\frac{1}{2\sigma_\mathrm{pde}^2} \sum_{j=1}^{N_c} \left[ r_1^2(x_j, y_j) + r_2^2(x_j, y_j) + r_3^2(x_j, y_j) \right],
    \label{eq:lik_pde}
\end{equation}
where $r_1$, $r_2$, and $r_3$ denote the $x$-momentum, $y$-momentum, and continuity residuals, respectively. All spatial derivatives are computed via automatic differentiation through the neural network in physical (non-normalized) coordinates.

\subsubsection{Tempered Posterior}
\label{sec:tempered_posterior}

A well-known challenge in Bayesian neural networks is that the standard posterior concentrates excessively as the number of data and collocation points grows, leading to overconfident uncertainty estimates~\cite{wenzel2020good}. We employ a tempered likelihood in which the effective sample size for each likelihood component is reduced from $N$ to $N^{\beta}$, where $\beta \in (0, 1)$ is a tempering exponent. The tempered log-posterior takes the form
\begin{equation}
    \log \tilde{p}(\theta \mid \mathcal{D}) = \log p(\theta) + N_d^{\beta_d} \, \overline{\log p_{UV}} + N_d^{\beta_f} \, \overline{\log p_f} + N_c^{\beta_r} \, \overline{\log p_r},
    \label{eq:tempered_posterior}
\end{equation}
where the overline denotes the mean (rather than sum) of the per-point log-likelihoods, and separate exponents $\beta_d$, $\beta_f$, and $\beta_r$ are assigned to the velocity data, Reynolds stress data, and physics residuals, respectively.

\paragraph{Justification of tempering exponents}
The choice of per-component tempering exponents $\beta_d$, $\beta_f$, and $\beta_r$ is motivated by both theoretical considerations and empirical calibration. From a theoretical perspective, the Bernstein--von Mises theorem guarantees that, for well-specified models with $N$ independent observations, the posterior concentrates at rate $\mathcal{O}(1/\sqrt{N})$. However, overparameterized neural networks violate the regularity conditions underlying this result: the parameters are highly correlated, the model is non-identifiable, and the effective number of degrees of freedom is substantially smaller than the number of observations~\cite{wenzel2020good}. As shown by Wenzel et al.~\cite{wenzel2020good}, tempering the likelihood by a factor $T < 1$ (equivalently, replacing $N$ with $N^\beta$ where $\beta < 1$) yields better-calibrated posteriors for deep neural networks. This phenomenon, known as the \emph{cold posterior effect}, has been widely observed across architectures and datasets~\cite{aitchison2021statistical,kapoor2022uncertainty}.

In the physics-informed setting, an additional source of miscalibration arises from the heterogeneous nature of the likelihood. The velocity data, Reynolds stress data, and PDE residuals contribute fundamentally different types of information: direct observations at sparse points versus soft constraints over the entire domain. Using the same effective sample size for all components would over-weight the PDE residuals (which are evaluated at $N_c \gg N_d$ collocation points) relative to the data, collapsing the posterior around solutions that satisfy the physics exactly at the expense of data fidelity. Per-component exponents $\beta_d > \beta_f > \beta_r$ address this imbalance by assigning progressively weaker influence to components with greater model-form uncertainty---the RANS closure terms $f_x$, $f_y$ and the PDE residuals carry inherent approximation errors that the Bayesian framework should not be forced to absorb entirely.

In practice, the exponents are selected via the following calibration procedure:
\begin{enumerate}
    \item An initial set of exponents is chosen based on the heuristic $\beta \approx 0.5$--$0.7$, which has been found to work well for Bayesian neural networks in regression tasks~\cite{wenzel2020good}.
    \item The NUTS sampler is run and the empirical coverage probability $C_q = \mathbb{P}(|q_\mathrm{true} - \hat{q}| < 2\hat{\sigma}_q)$ is computed for each output variable $q$ on a held-out evaluation set.
    \item The exponents are adjusted following the rule: if $C_q < 0.90$ (undercoverage), $\beta$ is decreased by $0.05$; if $C_q > 0.98$ (overcoverage) or the posterior mean RMSE degrades significantly relative to the MAP estimate, $\beta$ is increased by $0.05$.
    \item Steps 2--3 are repeated until the raw posterior coverage falls within $70$--$85\%$ for all variables, after which the post-hoc recalibration (Section~\ref{sec:recalibration}) brings the coverage to the nominal $95\%$ level.
\end{enumerate}
This procedure typically converges within 3--5 iterations. The final values $\beta_d = 0.70$, $\beta_f = 0.60$, $\beta_r = 0.50$ reflect the ordering of constraint reliability: the velocity data are the most trustworthy (highest $\beta$), the Reynolds stress observations carry moderate uncertainty (intermediate $\beta$), and the PDE residuals---which embed the RANS modeling approximation---are the least reliable as hard constraints (lowest $\beta$).

We note that the tempering exponents need not be tuned to high precision. The post-hoc recalibration procedure described in Section~\ref{sec:recalibration} corrects the global scale of the uncertainty estimates, so the primary role of the exponents is to ensure that: (i) the posterior mean remains accurate (close to the MAP estimate), and (ii) the \emph{spatial structure} of the posterior variance is physically meaningful, i.e., regions of higher prediction error correspond to regions of higher predicted uncertainty. As long as these two conditions are met, the recalibration step produces well-calibrated uncertainty intervals regardless of the exact exponent values.

\subsubsection{Two-Stage Inference Procedure}
\label{sec:two_stage}

Sampling from the tempered posterior is performed in two stages.

\paragraph{Stage 1: MAP Pre-training.}
To initialize the Markov chain in a high-probability region of the parameter space, we first obtain a maximum a posteriori (MAP) estimate $\theta_\mathrm{MAP}$ by minimizing the negative log-posterior using gradient-based optimization. This is performed in three sub-stages:
\begin{enumerate}[label=(\alph*)]
    \item \textit{Data-only optimization} (5,000 iterations with Adam~\cite{kingma2015adam}): the network is trained to fit the observed velocity and Reynolds stress fields without physics constraints, establishing an initial approximation of the data manifold.
    \item \textit{Data and physics with gradual ramp-up} (10,000 iterations with Adam): the physics residual loss is introduced with a weight that increases from zero to its full value following a cosine annealing schedule $\alpha(t) = \tfrac{1}{2}(1 - \cos(\pi t / T_w))$ over the first half of this sub-stage. The PDE residuals are evaluated on random mini-batches of 2,000 collocation points per iteration to manage computational cost.
    \item \textit{Second-order polishing} (5,000 L-BFGS iterations): the full tempered loss is minimized using the L-BFGS algorithm~\cite{liu1989limited} with strong Wolfe line search conditions to refine the MAP estimate near the optimum.
\end{enumerate}

\paragraph{Stage 2: NUTS Sampling}
Starting from $\theta_\mathrm{MAP}$, we draw samples from the tempered posterior using the No-U-Turn Sampler (NUTS)~\cite{hoffman2014no}, an adaptive variant of Hamiltonian Monte Carlo that automatically tunes the trajectory length. The sampler is configured with a maximum tree depth of 6, a target acceptance probability of 0.65, and 500 warmup iterations during which both the step size and a diagonal mass matrix are adapted. After warmup, $S = 500$ posterior samples are collected. To maintain tractable per-sample computational cost, the physics residual likelihood during NUTS is evaluated on a fixed random subsample of 1,000 collocation points, with the log-likelihood scaled by a factor of $N_c / 1{,}000$ to approximate the contribution of the full collocation set.

% ------------------------------------------------------------------
\subsection{MC Dropout Physics-Informed Neural Network}
\label{sec:mc_dropout}

Monte Carlo (MC) Dropout provides an approximate Bayesian inference framework
for neural networks by interpreting dropout at test time as a variational
approximation to the posterior distribution over the network
weights~\cite{gal2016dropout, zhang2019quantifying}. During training, a standard feedforward neural
network is optimized with dropout layers inserted after each hidden layer; at
inference time, dropout remains active and the forward pass is repeated
$S$~times with independently sampled dropout masks, yielding an ensemble of
stochastic predictions $\{\hat{u}^{(s)}(t)\}_{s=1}^{S}$ from which pointwise
predictive statistics are extracted:
\begin{equation}
  \bar{u}(t) = \frac{1}{S}\sum_{s=1}^{S} \hat{u}^{(s)}(t), \qquad
  \sigma^2(t) = \frac{1}{S}\sum_{s=1}^{S}
    \bigl(\hat{u}^{(s)}(t) - \bar{u}(t)\bigr)^2.
\end{equation}

In the physics-informed setting, the network is trained by minimizing a
composite loss comprising a data-fidelity term and a PDE residual term:
\begin{equation}\label{eq:mc_loss}
  \mathcal{L}(\boldsymbol{\theta})
  = \frac{1}{N_d}\sum_{i=1}^{N_d}
      \bigl(u_{\boldsymbol{\theta}}(t_i) - u_i^{\text{obs}}\bigr)^2
  + \lambda_{\text{pde}}\,\frac{1}{N_f}\sum_{j=1}^{N_f}
      \bigl[\mathcal{R}[u_{\boldsymbol{\theta}}](t_j)\bigr]^2,
\end{equation}
where $\{(t_i, u_i^{\text{obs}})\}_{i=1}^{N_d}$ are sparse noisy
observations, $\{t_j\}_{j=1}^{N_f}$ are collocation points, and
$\mathcal{R}[\cdot]$ denotes the PDE residual operator.

Compared to full Bayesian approaches such as Hamiltonian Monte Carlo
(HMC), MC~Dropout offers substantially lower computational cost since it
requires only a single training run followed by repeated forward passes,
rather than expensive MCMC sampling over the full weight space. However,
the quality of the resulting uncertainty estimates is inherently limited
by the expressiveness of the variational family induced by the dropout
distribution, which corresponds to a mixture of Bernoulli-masked weight
matrices \cite{psaros2023uncertainty}. In practice, this tends to produce well-calibrated uncertainty
in data-rich regions but may underestimate epistemic uncertainty in
extrapolation or data-sparse regimes, where the dropout masks do not
explore sufficiently diverse functional modes.

% ------------------------------------------------------------------
\subsection{Repulsive Deep Ensemble PINNs}
\label{sec:rde}

Ensemble methods have become a cornerstone in deep learning due to their conceptual simplicity, strong predictive performance, and inherent scalability \cite{lakshminarayanan2017simple, psaros2023uncertainty, zou2025learning, zou2024correcting}. Despite their success, maintaining functional diversity among independently trained ensemble members remains a fundamental challenge. In the absence of explicit mechanisms to enforce diversity, increasing the ensemble size often yields diminishing returns, with overall performance saturating near that of a single model. This limitation not only constrains predictive accuracy but also reduces the reliability of uncertainty estimates, particularly for out-of-distribution inputs.

Recent work by \cite{d2021repulsive} introduced Repulsive Ensembles, demonstrating that encouraging diversity among ensemble members can be formalized within a Bayesian framework. Specifically, they incorporate a kernel-based repulsive term into the ensemble update rules, which actively discourages different members from collapsing onto similar functional representations. This simple yet effective modification preserves ensemble diversity and enables the ensemble to approximate true Bayesian inference, in contrast to conventional deep ensembles that effectively perform only maximum a posteriori estimation. Furthermore, the training dynamics of repulsive ensembles can be interpreted as following a Wasserstein gradient flow of the KL divergence relative to the true posterior.

Building on this foundation, the present work adapts the repulsive ensemble methodology to PINNs. We examine the effects of repulsive interactions in both parameter and function space and provide comprehensive empirical comparisons with standard deep ensembles and Bayesian baselines. Our results focus on modeling mean flow and closure terms, demonstrating the benefits of maintaining functional diversity for both predictive accuracy and uncertainty quantification.

In what follows, we formulate the repulsive ensemble approach for PINNs in parameter and function space. In both cases, we consider an ensemble of $M$ neural networks parameterized by $\{\theta_1, \theta_2, \dots, \theta_M\}$, where $\theta_i \in \mathbb{R}^P$.

\subsubsection{Repulsive Deep Ensembles in Parameter Space}
\label{sec:rde_ps}

The parameter-space repulsive loss is defined as
\begin{equation}
\mathcal{L}_{\mathrm{rep}}^{\mathrm{param}}
=
\sum_{i<j}
\exp\!\left(
-\frac{\|\theta_i - \theta_j\|_2^2}{\sigma_\theta^2}
\right),
\end{equation}
where $\|\cdot\|_2$ denotes the Euclidean norm in parameter space, and $\sigma_\theta>0$ controls the repulsion strength. This term penalizes ensemble members with similar parameters to encourage diversity in weight space.

\subsubsection{Repulsive Deep Ensembles in Function Space}
\label{sec:rde_fs}

Let each network represent a function $f_{\theta_i} : \mathbb{R}^d \to \mathbb{R}^K$, and let $x \sim \mathcal{D}$ denote inputs sampled from the domain (e.g., collocation points). The function-space repulsive loss is
\begin{equation}
\mathcal{L}_{\mathrm{rep}}^{\mathrm{func}}
=
\sum_{i<j}
\mathbb{E}_{x \sim \mathcal{D}}
\left[
\exp\!\left(
-\frac{\|f_{\theta_i}(x) - f_{\theta_j}(x)\|_2^2}{\sigma_f^2}
\right)
\right],
\end{equation}
where $\|\cdot\|_2$ is the Euclidean norm in output space, and $\sigma_f>0$ sets the functional separation scale. For a discrete set of $N$ collocation points $\{x_n\}_{n=1}^N$, this is approximated by
\begin{equation}
\mathcal{L}_{\mathrm{rep}}^{\mathrm{func}}
\approx
\frac{1}{N}
\sum_{n=1}^{N}
\sum_{i<j}
\exp\!\left(
-\frac{\|f_{\theta_i}(x_n) - f_{\theta_j}(x_n)\|_2^2}{\sigma_f^2}
\right).
\end{equation}

For PINNs with vector outputs, e.g., $f(x) = (u,v,p,f_x,f_y) \in \mathbb{R}^5$, the functional norm is
\begin{equation}
\|f_{\theta_i}(x) - f_{\theta_j}(x)\|_2^2
=
\sum_{k=1}^{5}
\left(
f_{\theta_i}^{(k)}(x) - f_{\theta_j}^{(k)}(x)
\right)^2.
\end{equation}

\subsubsection{Full Training Objective}
\label{sec:rde_objective}

The total loss minimized by each ensemble member combines the PINN loss with the repulsive term:
\begin{equation}
\mathcal{L}
=
\mathcal{L}_{\mathrm{PINN}}
+
\lambda_{\mathrm{rep}}\,
\mathcal{L}_{\mathrm{rep}},
\end{equation}
where
\begin{equation}
\mathcal{L}_{\mathrm{PINN}}
=
\mathcal{L}_{\mathrm{data}}
+
\mathcal{L}_{\mathrm{PDE}},
\end{equation}
$\lambda_{\mathrm{rep}}$ controls the weight of the repulsion term, and $\mathcal{L}_{\mathrm{rep}}$ is either $\mathcal{L}_{\mathrm{rep}}^{\mathrm{param}}$ or $\mathcal{L}_{\mathrm{rep}}^{\mathrm{func}}$ depending on the chosen variant.

Unlike parameter-space repulsion, which enforces diversity in the weight space, function-space repulsion directly penalizes similarity in the outputs of ensemble members, resulting in physically meaningful ensemble diversity and improved epistemic uncertainty quantification for physics-informed neural networks.

% ------------------------------------------------------------------
\subsection{Predictive Uncertainty and Post-Hoc Recalibration}
\label{sec:recalibration}

Given the posterior samples $\{\theta^{(s)}\}_{s=1}^{S}$, pointwise predictive statistics are computed as
\begin{equation}
    \hat{q}(\mathbf{x}) = \frac{1}{S} \sum_{s=1}^{S} q_{\theta^{(s)}}(\mathbf{x}), \qquad \hat{\sigma}_q(\mathbf{x}) = \mathrm{std}\left( \{ q_{\theta^{(s)}}(\mathbf{x}) \}_{s=1}^{S} \right),
    \label{eq:predictive_stats}
\end{equation}
for each output quantity $q \in \{u, v, p, f_x, f_y\}$.

It is well established that Bayesian neural network posteriors---even those obtained via asymptotically exact methods such as HMC---tend to be miscalibrated, producing uncertainty estimates whose absolute scale does not match the observed prediction errors~\cite{kuleshov2018accurate,yao2019quality}. This miscalibration arises from several factors, including the use of tempered likelihoods, finite chain length, and the inherent difficulty of exploring high-dimensional, multimodal posterior landscapes.

To obtain well-calibrated uncertainty estimates, we apply a post-hoc recalibration procedure following the framework of Kuleshov et al.~\cite{kuleshov2018accurate}. For each output variable $q$, a scalar recalibration factor $\alpha_q$ is computed such that the rescaled uncertainty $\alpha_q \hat{\sigma}_q(\mathbf{x})$ achieves approximately 95\% empirical coverage. Specifically, we define
\begin{equation}
    \alpha_q = \frac{1}{2} \, P_{95}\!\left( \frac{|q_{\mathrm{true}}(\mathbf{x}_i) - \hat{q}(\mathbf{x}_i)|}{\hat{\sigma}_q(\mathbf{x}_i)} \right),
    \label{eq:alpha_calibration}
\end{equation}
where $P_{95}(\cdot)$ denotes the 95th percentile computed over the evaluation points. The calibrated uncertainty is then
\begin{equation}
    \sigma_q^\mathrm{cal}(\mathbf{x}) = \alpha_q \cdot \hat{\sigma}_q(\mathbf{x}).
    \label{eq:calibrated_sigma}
\end{equation}

This procedure preserves the spatial structure of the raw posterior uncertainty---regions identified as more uncertain by the NUTS posterior retain proportionally higher uncertainty after recalibration---while correcting the global scale to achieve nominal coverage. The recalibration factors $\alpha_q$ are computed on a held-out evaluation set and can be applied to predictions at new spatial locations without re-running the sampler.

For the pressure field $p$, for which no ground-truth data is available, the recalibration factor is taken as the average of $\alpha_u$ and $\alpha_v$, providing a physically motivated estimate based on the calibration quality of the velocity fields that govern the pressure through the momentum equations.

% ------------------------------------------------------------------

% ====================================================================
\section{Computational Experiments}
\label{sec:experiments}

% ------------------------------------------------------------------
\subsection{Van der Pol Oscillator: A Pedagogical Example}
\label{sec:vdp}

To demonstrate the efficiency of the proposed methods, we first consider the pedagogical problem of the Van der Pol oscillator. The governing equation is
\begin{align}
\frac{d^2 u}{d t^2}+\omega_0^2 u-\epsilon \omega_0\left(1-u^2\right) \frac{d u}{d t}=0,
\end{align}
with $\omega_0=15$, $\epsilon=1$, and a time horizon of $t=1.5$. The PDE residual operator for this problem is
\begin{equation}
  \mathcal{R}[u](t)
  = \ddot{u} - \varepsilon\,\omega\,(1 - u^2)\,\dot{u} + \omega^2\,u.
\end{equation}

The motivation behind choosing the Van der Pol oscillator is that its solution mimics the wake of the cylinder, and it has a limit cycle similar to  the flow past a cylinder.

\subsubsection{Van der Pol Oscillator: Bayesian PINN}
\label{sec:vdp_bpinn}
We solve the Van der Pol oscillator problem using the BPINNs method within the NeuralUQ framework \cite{zou2024neuraluq}.The underlying dynamical system is governed by a second-order nonlinear differential equation with parameters $\varepsilon = 1$ and $\omega = 15$, where the solution $x(t)$ is inferred from sparse noisy observations and physics constraints. The surrogate model is a fully connected neural network with architecture $[1, 50, 50, 1]$ and hyperbolic tangent activation functions, with all weights and biases assigned independent Gaussian priors $\mathcal{N}(0, 1)$. The training data consist of noisy observations obtained by subsampling the reference trajectory every $1500$ points and perturbing with Gaussian noise of standard deviation $\sigma = 0.05$, along with $N_f = 120$ uniformly distributed collocation points used to enforce the governing equation through a residual formulation scaled by $10^{-4}$ for numerical stability. Both data and physics constraints are incorporated via Gaussian likelihoods with fixed standard deviation $\sigma = 0.05$.

\begin{figure}
    \centering

    \begin{subfigure}[b]{0.46\linewidth}
        \centering
        \includegraphics[height=6cm]{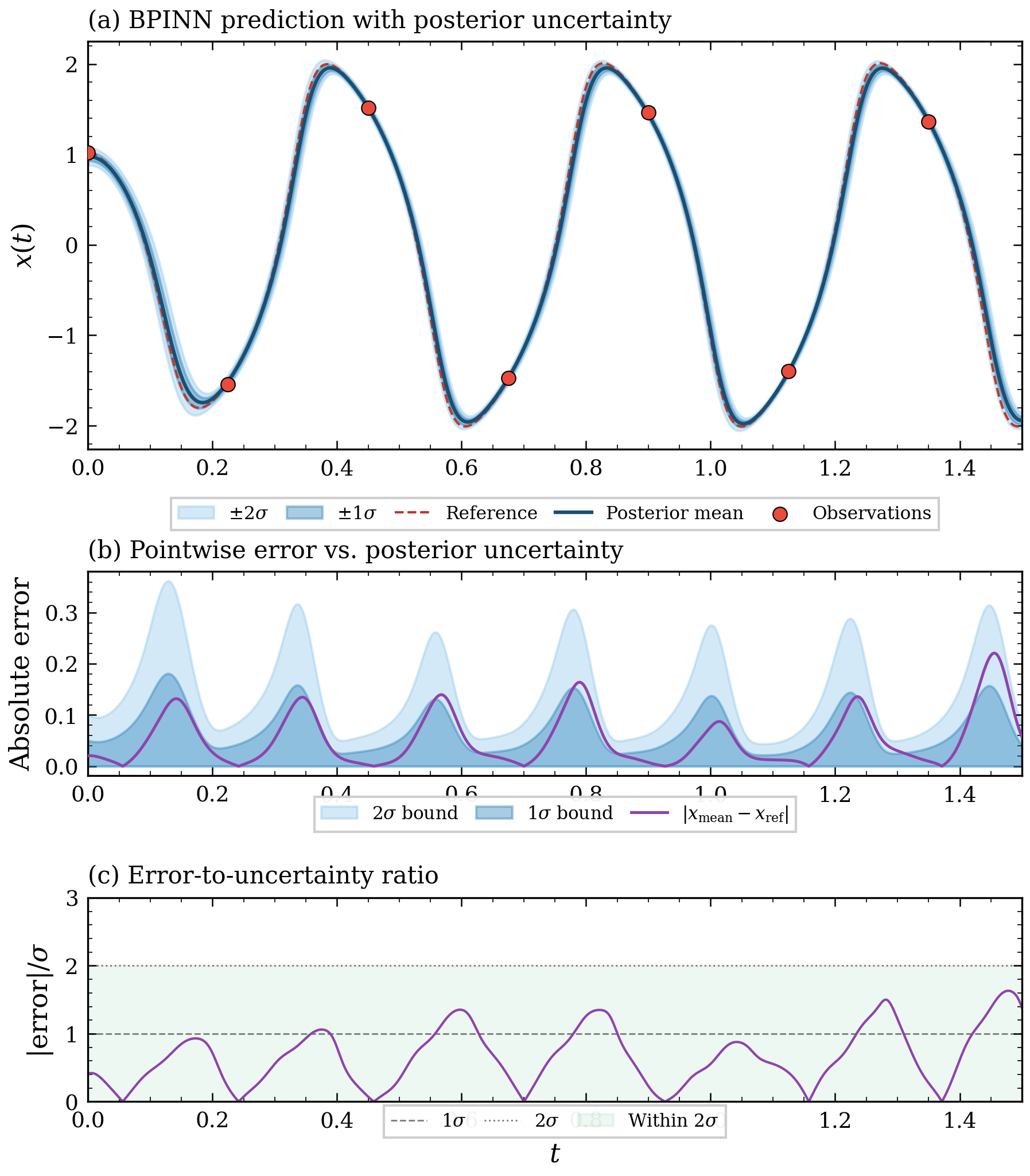}
        \caption{Bayesian PINN: uncertainty quantification versus prediction error.}
        \label{fig:vdp_bpinn}
    \end{subfigure}
    \hspace{0.01\linewidth}
    \begin{subfigure}[b]{0.40\linewidth}
        \centering
        \includegraphics[height=6cm]{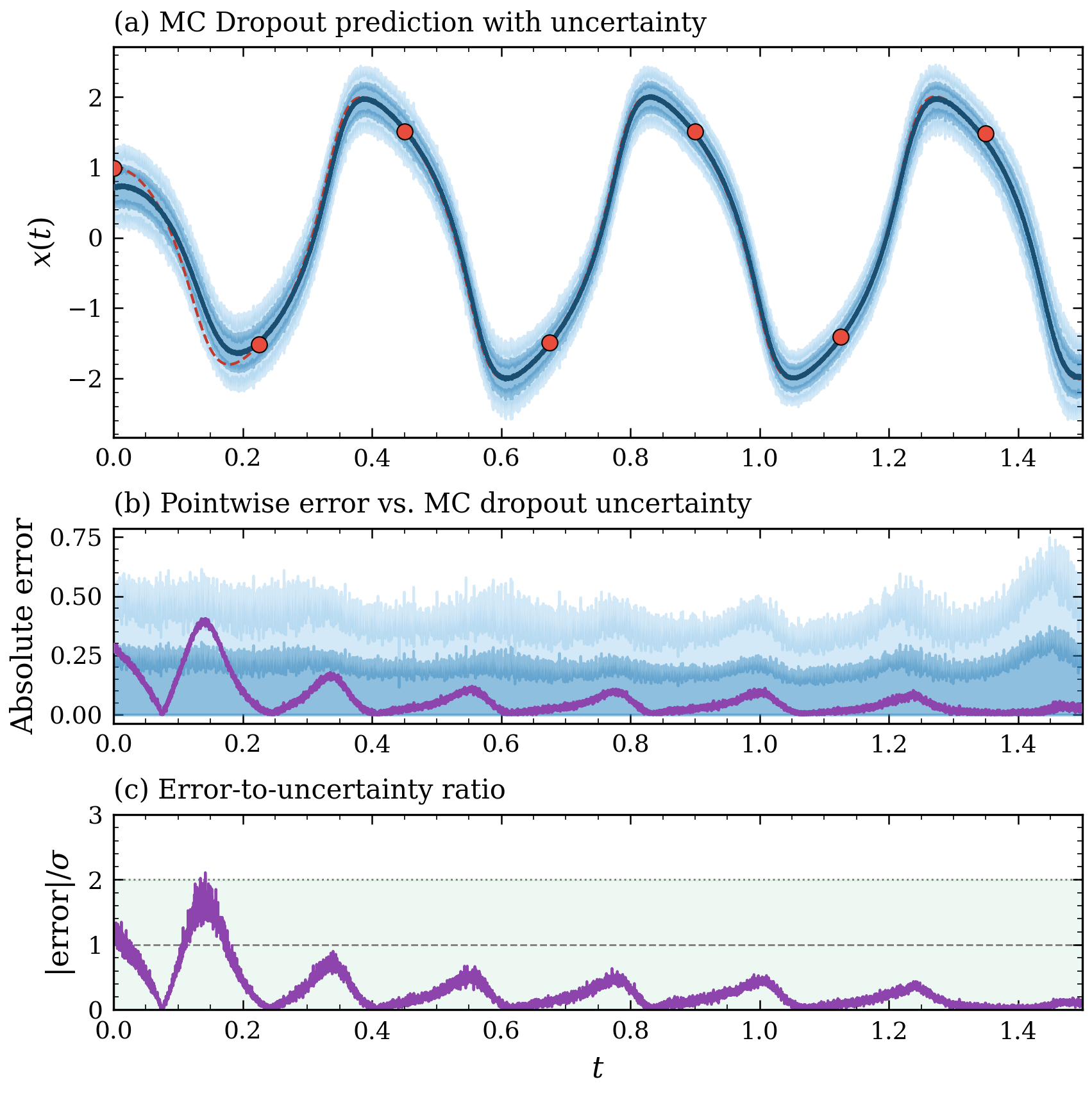}
        \caption{MC Dropout: uncertainty quantification versus prediction error.}
        \label{fig:vdp_mc_dropout}
    \end{subfigure}

    \caption{Van der Pol oscillator: comparison of uncertainty quantification obtained  versus prediction error using BPINN (left)  and MC Dropout (right).}
    \label{fig:vdp_comparison}
\end{figure}

Posterior inference is performed using Hamiltonian Monte Carlo (HMC) with $1000$ burn-in steps and $1000$ posterior samples, employing a step size of $0.01$ and $50$ leapfrog steps per iteration. The posterior predictive distribution is obtained by propagating the sampled network parameters through the model, yielding the predictive mean and standard deviation across the full temporal domain. In \autoref{fig:vdp_bpinn}, we present the posterior mean along with the calibration of uncertainty against the pointwise absolute error. The results demonstrate that the posterior mean closely matches the reference solution (top panel in \autoref{fig:vdp_bpinn}), even in regions with limited observational data, while the uncertainty estimates appropriately reflect model confidence. Quantitatively, the mean absolute error is small compared to the predicted uncertainty (middle panel in \autoref{fig:vdp_bpinn}), and the empirical coverage shows that approximately $68\%$ and $95\%$ of the errors lie within the $\pm 1\sigma$ and $\pm 2\sigma$ credible intervals, respectively (bottom panel in \autoref{fig:vdp_bpinn}), indicating well-calibrated uncertainty. Furthermore, the ratio of pointwise error to predicted standard deviation remains close to unity across most of the domain, confirming that the BPINN provides both accurate predictions and reliable uncertainty quantification.

\subsubsection{Van der Pol Oscillator: MC Dropout}
\label{sec:vdp_mcdropout}

For the Van der Pol oscillator, the MC Dropout PINN employs a fully connected neural network with four hidden layers of $48$ units each and $\tanh$ activation functions, with a dropout layer (rate $p = 0.001$) applied after every hidden layer. The relatively low dropout rate is chosen to preserve deterministic predictive accuracy while introducing sufficient stochasticity to enable uncertainty quantification at test time. The model is trained for $2 \times 10^5$ epochs using the AdamW optimizer with a learning rate of $10^{-4}$. During inference, $S = 1000$ stochastic forward passes are performed with dropout activated, yielding an approximate posterior predictive distribution.

The results in \autoref{fig:vdp_mc_dropout} indicate that, in comparison to the BPINN approach, MC Dropout produces less reliable uncertainty estimates. In particular, the method tends to be overly conservative, overestimating uncertainty even in regions where observational data are available (top panel in \autoref{fig:vdp_mc_dropout}). This leads to poorer calibration, as the predicted uncertainty does not consistently reflect the true pointwise error (middle and bottom panel in \autoref{fig:vdp_mc_dropout}. Consequently, while MC Dropout provides a computationally efficient approximation, it lacks the probabilistic consistency and calibration accuracy achieved by the fully Bayesian BPINN framework.

\begin{figure}[htbp]
    \centering
    \begin{subfigure}[b]{0.48\linewidth}
        \centering
        \includegraphics[width=\linewidth]{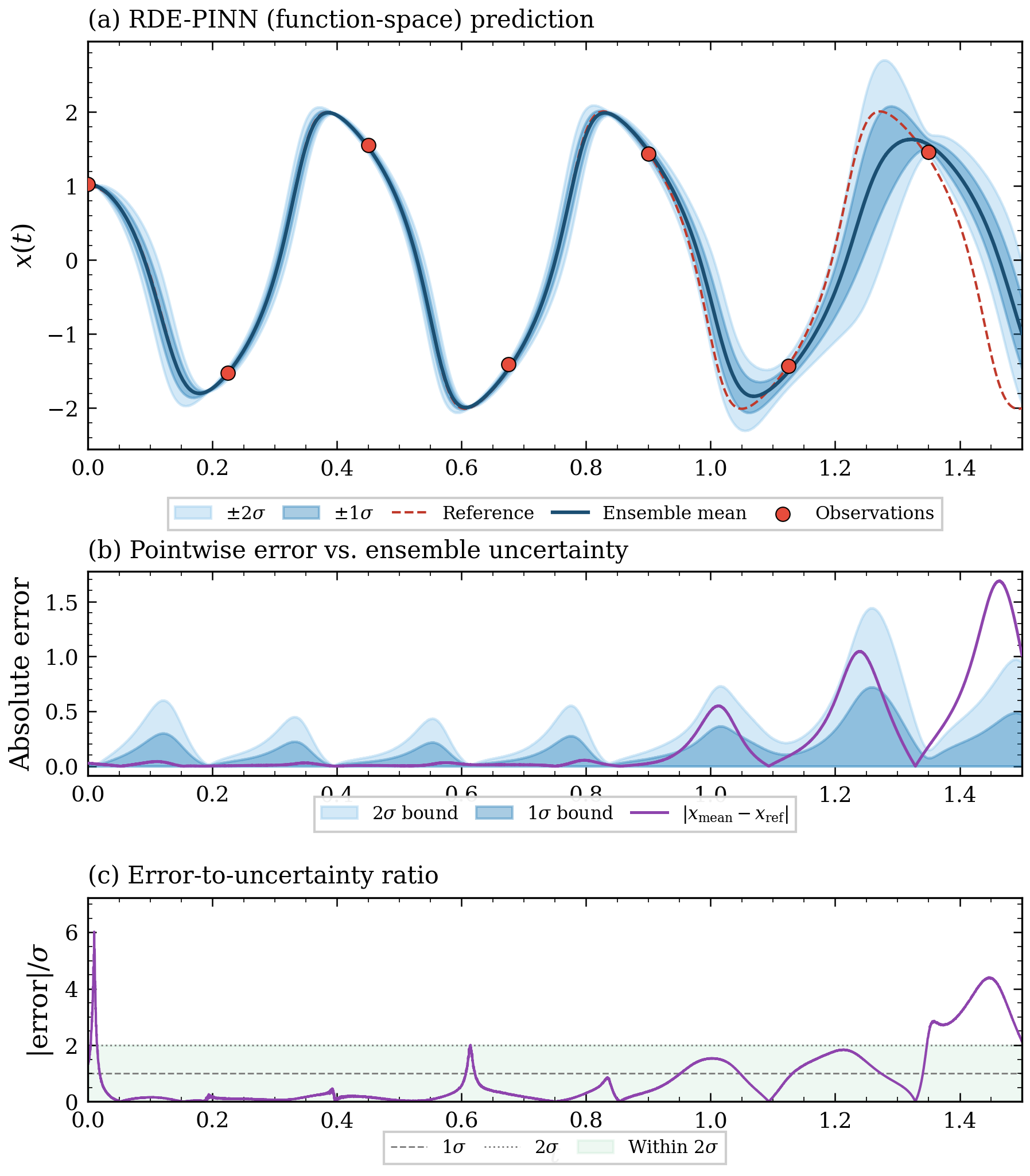}
        \caption{Repulsive Deep Ensembles (function space): uncertainty quantification versus prediction error.}
        \label{fig:vdp_rde_fs}
    \end{subfigure}
    \hfill
    \begin{subfigure}[b]{0.48\linewidth}
        \centering
        \includegraphics[width=\linewidth]{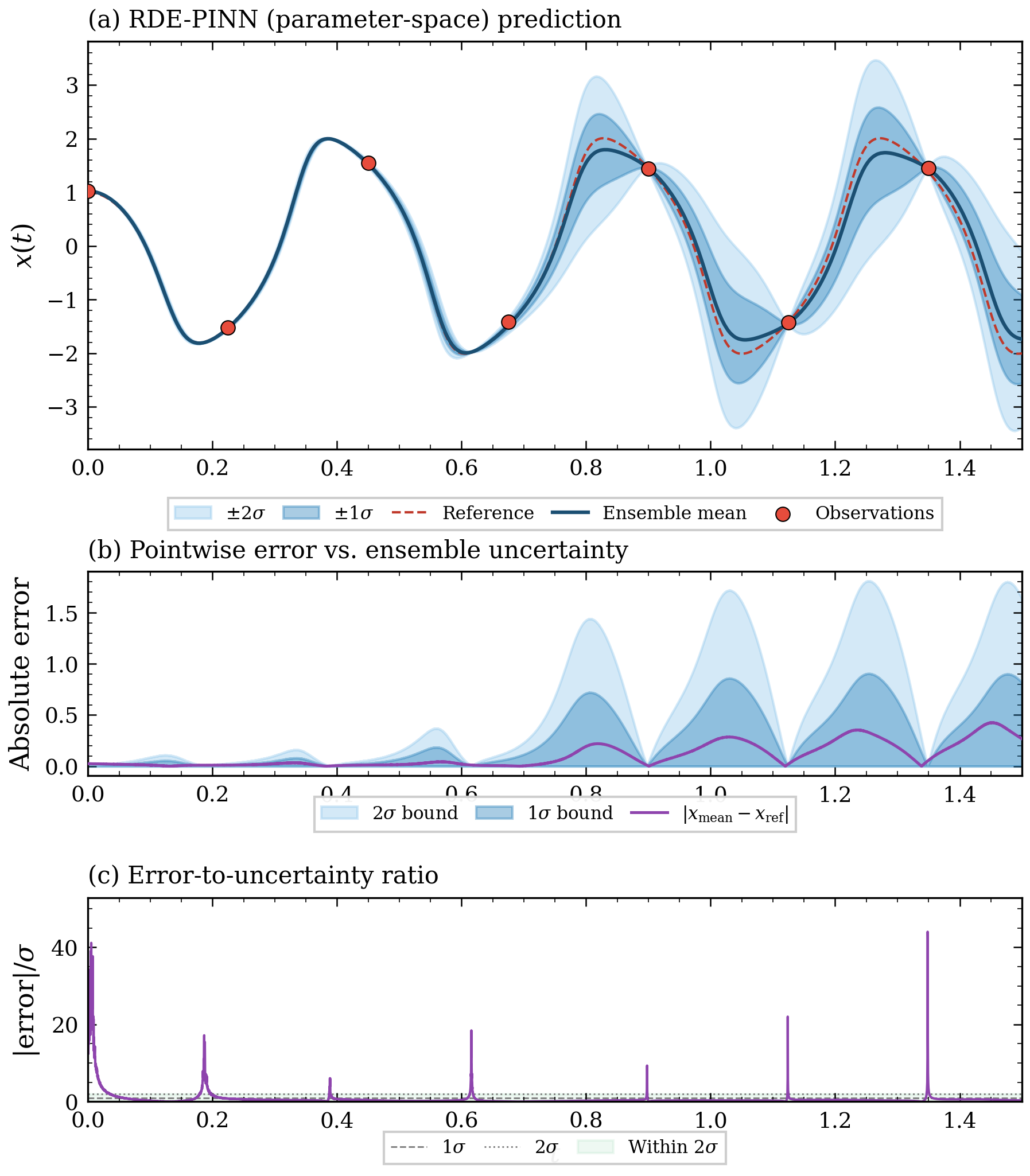}
        \caption{Repulsive Deep Ensembles (parameter space): uncertainty quantification versus prediction error.}
        \label{fig:vdp_rde_ps}
    \end{subfigure}

    \caption{Van der Pol oscillator: comparison of Repulsive Deep Ensembles in parameter space and function space for uncertainty quantification versus prediction error.}
    \label{fig:vdp_rde_comparison}
\end{figure}

\subsubsection{Van der Pol Oscillator: Repulsive Deep Ensembles in function and parameter Space}
\label{sec:vdp_rde_fs}
We consider the Van der Pol oscillator using a Repulsive Deep Ensemble physics-informed neural network (RDE-PINN), where diversity among ensemble members is explicitly encouraged through a repulsive regularization defined in function space. The model consists of an ensemble of $M = 10$ neural networks, each with depth $5$ and width $32$, employing $\tanh$ activation functions. The training objective combines a data misfit term, a physics-informed residual enforcing the governing differential equation, and a repulsive loss defined over the predicted functions. Specifically, $N_f = 200$ collocation points are used to enforce the differential equation, while sparse noisy observations are obtained by subsampling the reference solution every $1500$ points and perturbing them with Gaussian noise of standard deviation $0.05$. An additional set of $N_{\mathrm{rep}} = 64$ points is used to compute the functional repulsion term, which penalizes similarity between ensemble predictions via a kernel-based measure. The model is trained for $3 \times 10^5$ epochs using the AdamW optimizer with a cosine decay learning rate schedule initialized at $10^{-4}$, while the contribution of the physics loss is gradually increased during training to balance data fidelity and physical consistency.

The results in \autoref{fig:vdp_rde_fs} demonstrate that enforcing diversity in function space leads to improved predictive accuracy and well-calibrated uncertainty estimates. The ensemble mean closely matches the reference solution (top panel), while the predictive uncertainty adapts to regions of higher model discrepancy (middle and bottom panels). Quantitatively, the ratio of the pointwise absolute error to the predicted standard deviation remains close to unity across most of the domain, and the empirical coverage within the $\pm 1\sigma$ and $\pm 2\sigma$ intervals is consistent with the expected $68\%$ and $95\%$ levels, indicating reliable calibration. Furthermore, the pairwise functional distances between ensemble members remain sufficiently large, confirming that the repulsive mechanism effectively promotes meaningful diversity in the predicted solutions.

For comparison, we also consider a parameter-space RDE-PINN, where the repulsive regularization is applied to the network weights rather than the model outputs. The architecture, data, and training procedure remain identical, with $M = 10$ ensemble members, depth $5$, width $32$, $N_f = 200$ collocation points, and training over $3 \times 10^5$ epochs using AdamW with a cosine decay schedule. In this case, the repulsive term penalizes similarity between flattened parameter vectors, encouraging the ensemble members to occupy different regions of the weight space. As in the function-space formulation, the physics residual is gradually introduced during training through a continuation strategy.

The results in \autoref{fig:vdp_rde_ps} show that parameter-space repulsion leads to less reliable uncertainty estimates. In particular, the predicted uncertainty tends to be underestimated relative to the true error, as observed in the top and middle panels, and the error-to-uncertainty ratio is significantly larger than unity compared to the function-space approach. This indicates poor calibration, with the predicted uncertainty failing to adequately reflect the actual prediction error.

Comparing \autoref{fig:vdp_rde_fs} and \autoref{fig:vdp_rde_ps}, the function-space RDE-PINN clearly provides more meaningful and reliable uncertainty quantification. While both approaches encourage diversity among ensemble members, repulsion in function space acts directly on the predicted solutions, promoting variability that is consistent with the underlying dynamics. In contrast, parameter-space repulsion only separates network weights, which does not guarantee sufficiently distinct functional representations, since different parameter configurations can produce similar outputs.

As a result, the function-space formulation achieves better alignment between uncertainty and pointwise error, improved calibration of predictive intervals, and superior overall predictive performance. These results highlight that, for physics-informed ensemble models, enforcing diversity in function space is more effective than doing so in parameter space when the goal is accurate and trustworthy uncertainty quantification.

We also conducted experiments using a vanilla deep ensemble approach \cite{psaros2023uncertainty}. However, this method failed to produce meaningful uncertainty estimates—the predictions collapsed, and the reported variance was effectively zero \ref{app:vpol_dens}.

\paragraph{Takeaway from the Van der Pol oscillator}
The detailed uncertainty quantification (UQ) analysis for the Van der Pol oscillator shows that the BPINN framework delivers the most reliable performance, primarily due to its fully Bayesian formulation and well-calibrated uncertainty estimates. Repulsive Deep Ensembles (RDE) provide a computationally efficient approximation to Bayesian inference; however, their effectiveness depends critically on how diversity is enforced. In particular, the function-space RDE produces meaningful and well-calibrated uncertainty, whereas the parameter-space RDE leads to poorly calibrated estimates, since diversity in network weights does not consistently translate into diversity in the predicted solutions. Additionally, vanilla deep ensembles are not well-suited for this problem, as they result in a collapse of predictions, with all ensemble members converging to nearly identical outputs. In contrast, the MC Dropout-based approach performs significantly worse, as it tends to overestimate uncertainty in regions where observational data are available, leading to overly conservative and less informative predictions. 

Based on these observations, we focus our subsequent analysis on BPINN and function-space RDE methods. While parameter-space RDE is conceptually appealing, it becomes computationally expensive and impractical for large-scale and ill-posed problems such as the RANS equations. In particular, enforcing repulsion in parameter space requires pairwise comparisons between ensemble members, leading to $\mathcal{O}(M^2)$ complexity, which becomes prohibitive for large networks and ensemble sizes. Consequently, parameter-space RDE is not pursued further in this work.

% ------------------------------------------------------------------
\subsection{Flow Past a Cylinder}
\label{sec:cylinder}

We apply the uncertainty quantification methods described in Section~\ref{sec:methodology} to the RANS inverse problem defined by Eqs.~\eqref{eq:mom_x}--\eqref{eq:continuity}. For the cylinder flow problems, the network architecture consists of five hidden layers of 64 neurons each with hyperbolic tangent activation functions, yielding approximately 17,000 trainable parameters. We consider two Reynolds numbers: $\mathrm{Re} = 3{,}900$ using DNS reference data, and $\mathrm{Re} = 10{,}000$ using experimental measurements.

% ------------------------------------------------------------------
\subsubsection{$\mathrm{Re} = 3{,}900$: DNS Data}
\label{sec:re3900}
\paragraph{Data generation and data used for UQ analysis}
Incompressible flow past a circular cylinder at Reynolds number $\mathrm{Re}=3900$ is simulated using the open-source spectral element solver \texttt{nekRS}~\cite{fischer2022nekrs}. Direct numerical simulation (DNS) is employed without any turbulence modeling to obtain high-fidelity data. The computational domain is $[-7.5D, 25D] \times [-10D, 10D]$ in the streamwise ($x$) and crossflow ($y$) directions, with the cylinder centered at $(0,0)$. The discretization uses a spectral element method of order $7$, with $N_{xy}=1607$ elements in the $x$--$y$ plane, $N_c=96$ elements along the cylinder circumference, and $N_z=64$ elements in the spanwise direction ($L_z/D=6.4$), resulting in approximately $35.3$ million degrees of freedom. Instantaneous flow fields are time-averaged to obtain mean velocity and pressure, and Reynolds stresses are computed from the corresponding velocity fluctuations.

\begin{figure}
    \centering
    \includegraphics[width=\linewidth]{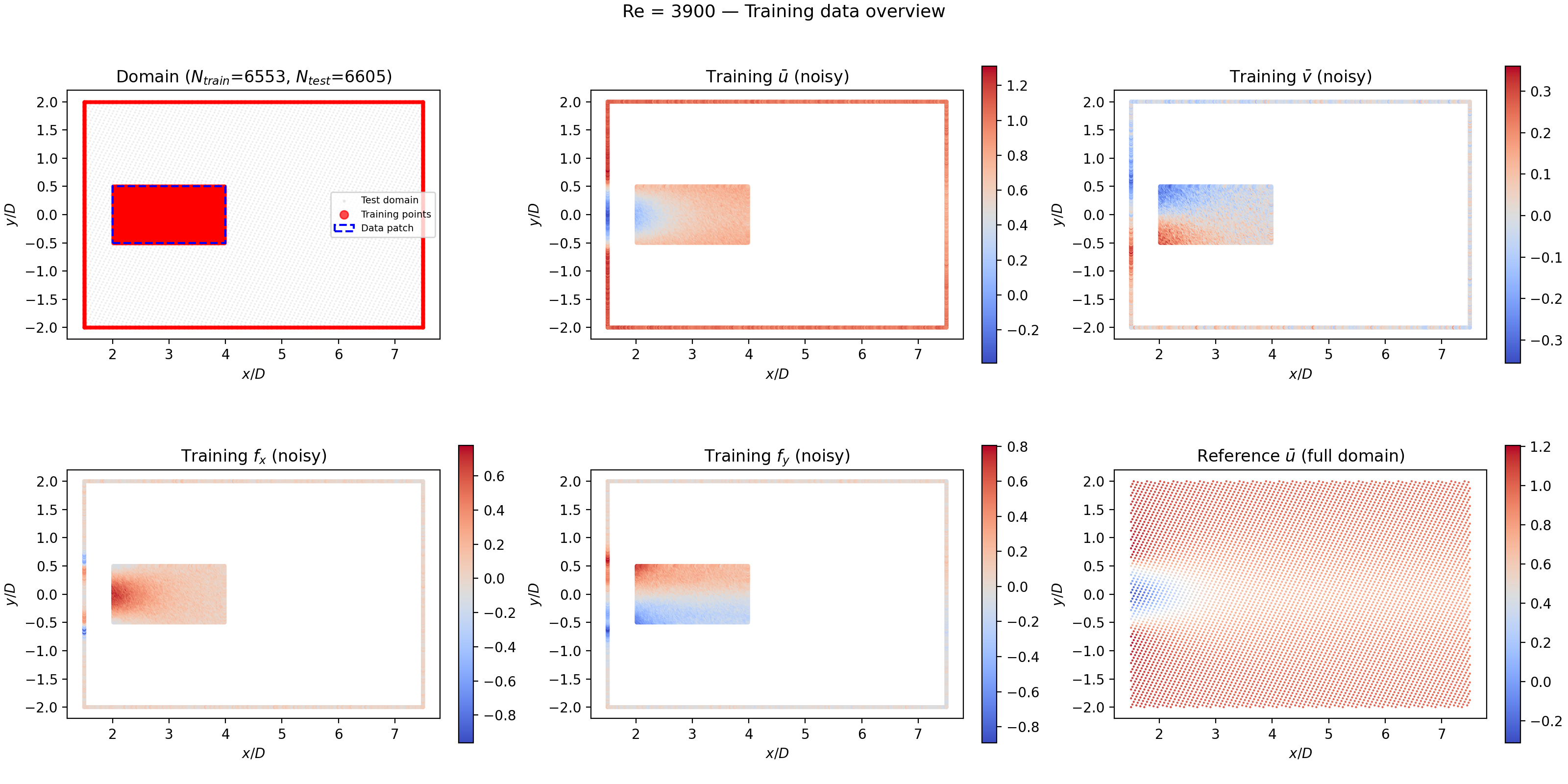}
    \caption{$\mathrm{Re}=3{,}900$: Computational domain and data used for training and inference. The top-left panel shows the spatial distribution of sparse observation points used for Bayesian inference across the domain. The remaining panels in the top row illustrate the perturbed velocity measurements, where Gaussian noise is added to the underlying data. The first two panels in the bottom row show the Reynolds stress components with additive Gaussian perturbations. The final panel presents the distribution of test points used for inference, which densely cover the computational domain.}
    \label{fig:re3900_domain_training}
\end{figure}

The resulting dataset is used for uncertainty quantification (UQ), where the objective is to infer flow quantities from sparse and noisy observations. As shown in \autoref{fig:re3900_domain_training}, only a limited number of observation points are used, with Gaussian noise added to both velocity and Reynolds stress data to mimic realistic measurement uncertainty. A separate set of collocation points enforces the governing equations, while a dense set of test points is used for evaluation. This setup provides a challenging benchmark for assessing the accuracy and calibration of the proposed UQ methods under data-scarce conditions.

\paragraph{Bayesian PINN}
We consider uncertainty quantification for incompressible turbulent flow past a circular cylinder at $\mathrm{Re}=3900$ using a Bayesian physics-informed neural network (BPINN). The surrogate model is a fully connected neural network with architecture $[2, 64, 64, 64, 64, 64, 5]$, where the outputs correspond to the mean velocity components $(U, V)$, pressure $\bar{p}$, and closure terms $(f_x, f_y)$. The governing Reynolds-averaged Navier--Stokes (RANS) equations are enforced through a physics-informed residual formulation, where the closure terms are learned directly as network outputs, enabling a data-driven representation of unresolved stresses. Training data consist of sparse and noisy observations of $(U, V, f_x, f_y)$, with additive Gaussian noise of standard deviation $0.05$, while pressure is inferred purely through the governing equations without direct supervision. The PDE residuals are enforced at collocation points and modeled using a Gaussian likelihood with standard deviation $\sigma_{\mathrm{PDE}} = 0.3$.

\begin{figure}
    \centering
    \includegraphics[trim={0cm 7.9cm 0cm 0cm},clip,width=\linewidth]{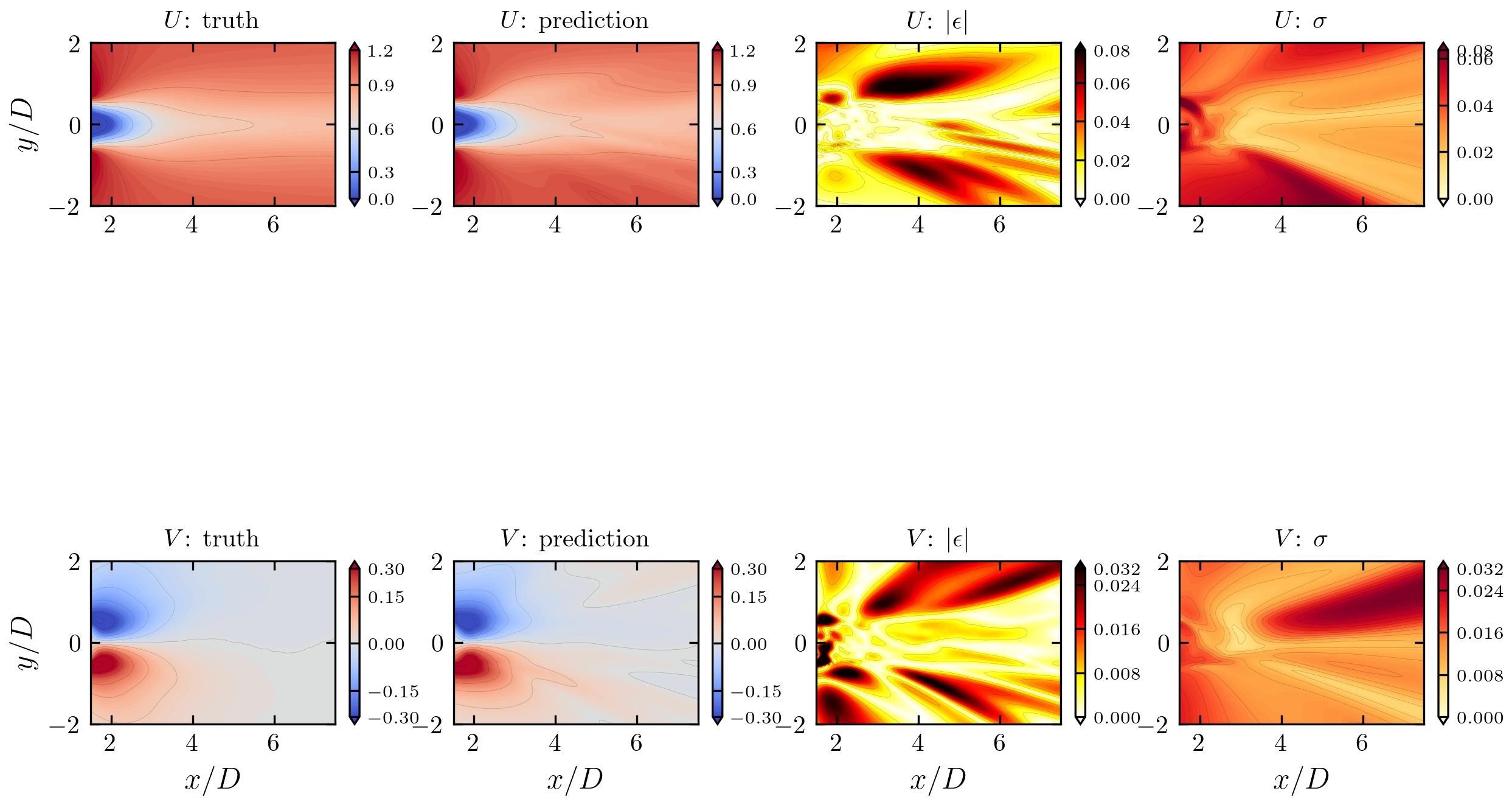}
    \includegraphics[trim={0cm 0cm 0cm 7cm},clip,width=\linewidth]{figs/Re_3900_Cyl/fig_velocity_diagnostics.png}
    \caption{$\mathrm{Re}=3{,}900$ --- Bayesian PINN velocity diagnostics. The panels show the reference solution, posterior mean prediction, absolute error, and predictive standard deviation, ratio for the velocity components $({U}, {V})$. The results demonstrate accurate reconstruction of the velocity field and good alignment between predicted uncertainty and pointwise error across the domain.}
    \label{fig:re3900_bpinn_vel}
\end{figure}

The training procedure follows a staged maximum a posteriori (MAP) pre-training strategy, consisting of a data-only phase, a gradual incorporation of physics constraints via a cosine ramp, and a final refinement using an LBFGS optimizer. To ensure computational efficiency, mini-batching is employed for evaluating PDE residuals. Subsequently, full Bayesian inference is performed using the No-U-Turn Sampler (NUTS) with $500$ warmup steps and $500$ posterior samples, with subsampling of collocation points to reduce computational cost. Independent Gaussian priors with variance $\sigma_{\mathrm{prior}}^2 = 4$ are placed over all network parameters, and likelihood tempering is used to balance the relative contributions of data and physics terms. The posterior predictive distribution is obtained by propagating sampled network parameters through the model, yielding estimates of the predictive mean and uncertainty for all flow variables. To improve calibration, a post-hoc scaling of the predictive standard deviation is applied based on empirical coverage. 

\begin{figure}
    \centering
    \includegraphics[trim={0cm 7.9cm 0cm 0cm},clip,width=\linewidth]{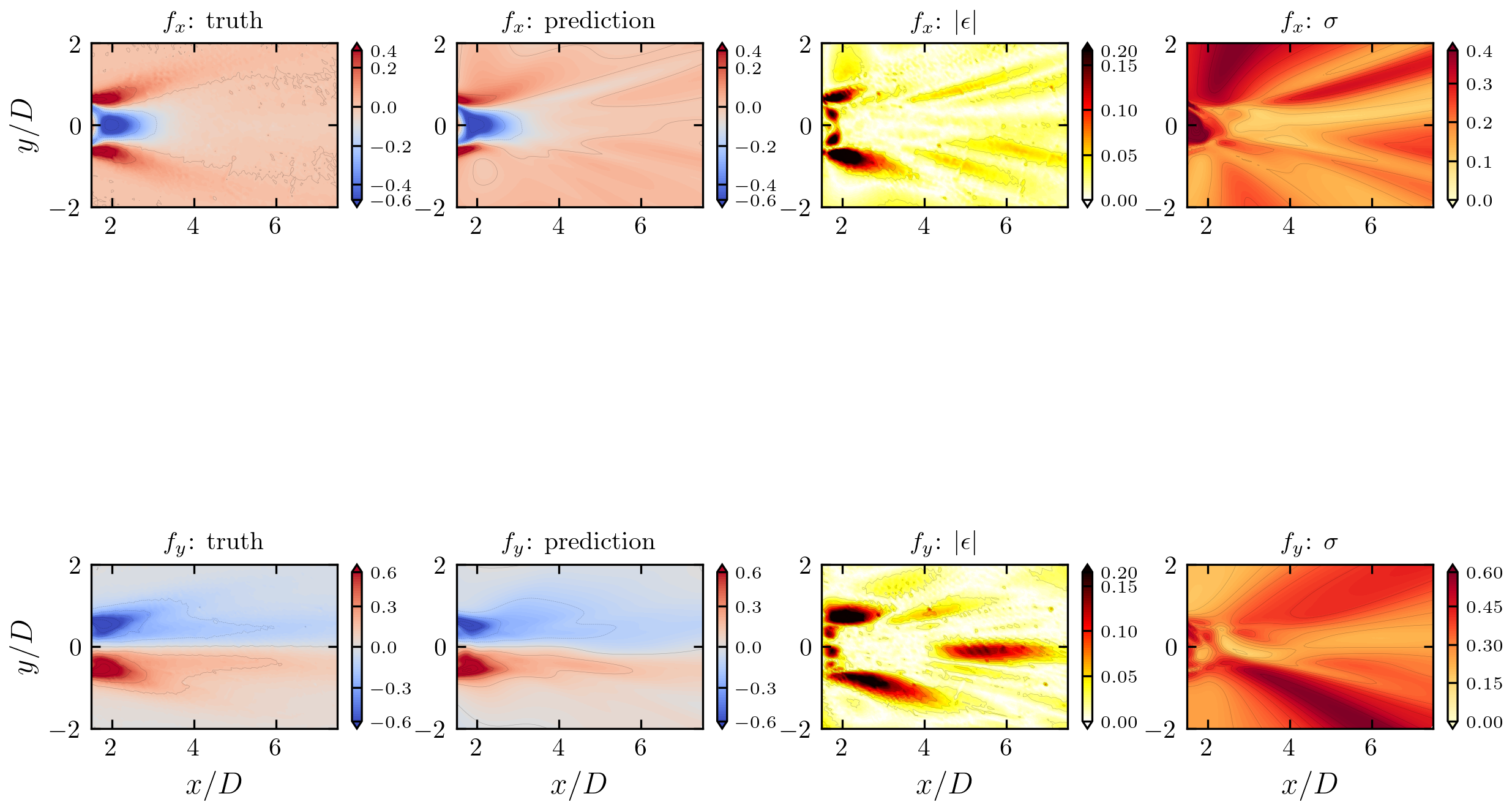}
    \includegraphics[trim={0cm 0cm 0cm 7cm},clip,width=\linewidth]{figs/Re_3900_Cyl/fig_reynolds_stress_diagnostics.png}
    \caption{$\mathrm{Re}=3{,}900$ --- Bayesian PINN Reynolds stress diagnostics. The panels present the reference fields, posterior mean predictions, absolute error, and predictive standard deviation for the Reynolds stress divergence components $(f_x, f_y)$. The results indicate accurate recovery of the closure terms and well-calibrated uncertainty estimates throughout the computational domain.}
    \label{fig:re3900_bpinn_rs}
\end{figure}

The results for the Bayesian PINN at $\mathrm{Re}=3{,}900$ demonstrate accurate prediction of both the velocity field and Reynolds stress divergence, along with well-calibrated uncertainty estimates after post-processing. As shown in \autoref{fig:re3900_bpinn_vel} and \autoref{fig:re3900_bpinn_rs}, the posterior mean closely matches the reference solution for the velocity components $(U, V)$ and the closure terms $(f_x, f_y)$, while the spatial distribution of predictive uncertainty captures regions of higher model discrepancy. The error-to-uncertainty ratio further indicates strong alignment between predicted uncertainty and pointwise error across the domain. Quantitative metrics reported in \autoref{tab:re3900_bpinn} confirm these observations. For the velocity and pressure fields, the RMSE remains low (e.g., $0.0303$ for $U$ and $0.0119$ for $V$), and the recalibrated uncertainty achieves the target coverage of approximately $95\%$ within the $\pm 2\sigma$ interval. 

Similarly, for the Reynolds force components, although the prediction errors are larger due to the increased complexity of the closure terms, the recalibrated uncertainties remain well calibrated, with coverage restored from severely underconfident raw estimates (as low as $2.5\%$ for $f_y$) to the nominal $95\%$ level. The mean error-to-uncertainty ratios are close to unity for all variables, indicating consistent and reliable uncertainty quantification. Overall, these results highlight the ability of the BPINN framework to provide accurate predictions and well-calibrated uncertainty estimates for both primary flow variables and derived closure quantities in turbulent flows. 

\begin{table}
  \centering
  \caption{Prediction accuracy and uncertainty calibration for the Bayesian PINN at $\mathrm{Re}=3{,}900$. $\bar{\sigma}_\mathrm{raw}$: raw posterior standard deviation; $\bar{\sigma}_\mathrm{cal}$: recalibrated standard deviation; $C_{2\sigma}$: coverage probability ($|\epsilon| < 2\sigma$); $\langle|\epsilon|/\sigma\rangle$: mean error-to-uncertainty ratio.}
  \label{tab:re3900_bpinn}
  \begin{tabular}{lcccccc}
    \toprule
    Variable & RMSE & $\bar{\sigma}_\mathrm{raw}$ & $\bar{\sigma}_\mathrm{cal}$ & $C_{2\sigma}^\mathrm{raw}$ & $C_{2\sigma}^\mathrm{cal}$ & $\langle|\epsilon|/\sigma_\mathrm{cal}\rangle$ \\
    \midrule
    $U$ & 0.0303 & 0.0058 & 0.0337 & 32.6\% & 95.0\% & 0.74 \\
    $V$ & 0.0119 & 0.0053 & 0.0143 & 62.6\% & 95.0\% & 0.73 \\
    $P$ & 0.0277 & 0.0064 & 0.0289 & 29.1\% & 95.0\% & 0.93 \\
    $f_x$ & 0.1913 & 0.0068 & 0.2060 & 19.9\% & 95.0\% & 0.42 \\
    $f_y$ & 0.3211 & 0.0077 & 0.3468 & 2.5\% & 95.0\% & 0.68 \\
    \bottomrule
  \end{tabular}
\end{table}

\begin{figure}
    \centering
    \includegraphics[width=\linewidth]{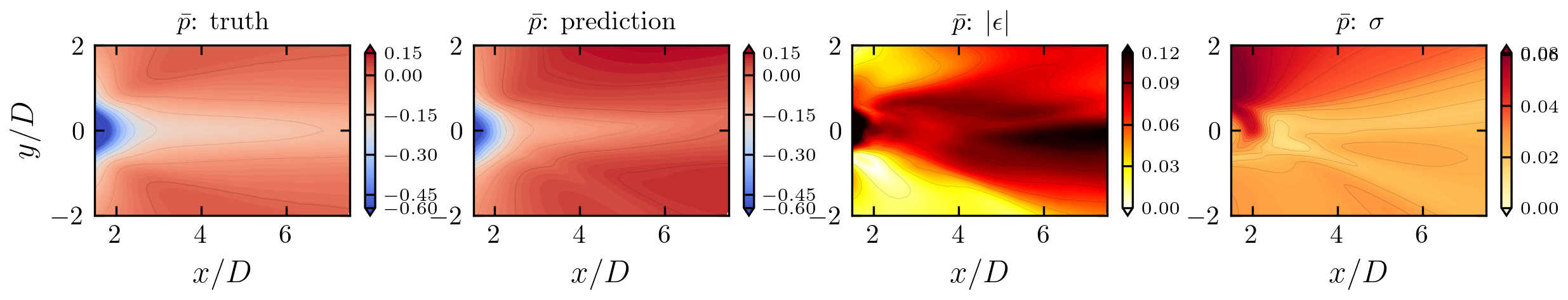}
    \caption{$\mathrm{Re}=3{,}900$ --- Bayesian PINN: inferred pressure field. No pressure data are used during training, and the pressure is recovered solely through the governing equations. The posterior mean prediction closely matches the reference solution, while the predictive uncertainty provides a meaningful estimate of the local error, which is particularly important in practical scenarios where reference pressure data are unavailable.}
    \label{fig:re3900_bpinn_pressure}
\end{figure}

The pressure field shown in \autoref{fig:re3900_bpinn_pressure} is inferred entirely from the governing equations without any direct supervision. Despite the absence of pressure data during training, the BPINN accurately reconstructs the pressure distribution across the domain. Moreover, the predictive uncertainty provides a reliable indication of the local error, offering a practical measure of confidence in regions where ground-truth pressure data are not available. This highlights a key advantage of the BPINN framework, namely its ability to recover latent quantities and provide physically consistent uncertainty estimates in data-scarce settings.

\paragraph{Repulsive Deep Ensembles in Function Space}
For turbulent flow past a circular cylinder at $\mathrm{Re}=3900$, we employ a Repulsive Deep Ensemble physics-informed neural network (RDE-PINN) with diversity enforced in function space. The model consists of an ensemble of $M=10$ neural networks, each with depth $5$ and width $64$, mapping the spatial coordinates $(x,y)$ to the outputs $(U, V, P, f_x, f_y)$. Training data consist of sparse noisy observations of $U$, $V$, $f_x$, and $f_y$, while no pressure data are used during training; instead, pressure is inferred entirely through the governing equations. The training set combines boundary data with an additional near-wake patch, and additive Gaussian noise of standard deviation $0.05$ is imposed on all observed quantities. To enforce the physics, $N_{\mathrm{col}}=2000$ collocation points are sampled in the spatial domain, while $N_{\mathrm{rep}}=200$ points are used to evaluate the repulsive loss. The total loss combines a data misfit term, a PDE residual term based on the incompressible Reynolds-averaged Navier--Stokes equations, and a function-space repulsive regularization term, which promotes diversity by penalizing similarity among ensemble predictions. The model is trained for $10^5$ epochs using the AdamW optimizer with an initial learning rate of $10^{-3}$ and a cosine decay schedule, while the PDE contribution is gradually ramped up during training.
\begin{figure}
    \centering
    \includegraphics[trim={0cm 7.9cm 0cm 0cm},clip,width=\linewidth]{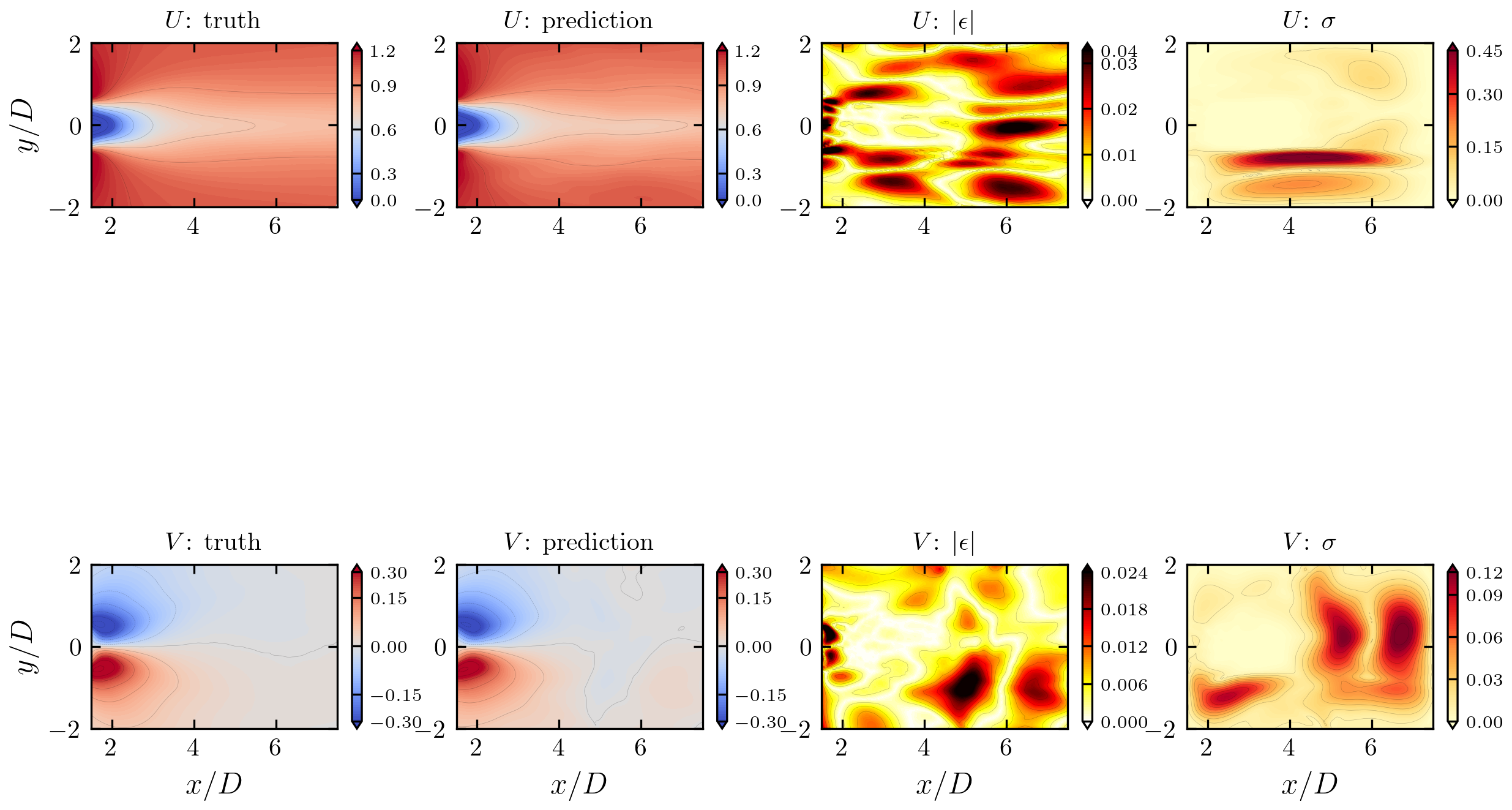}
    \includegraphics[trim={0cm 0cm 0cm 7cm},clip,width=\linewidth]{figs/RDE_Re_3900_FS/fig_velocity_diagnostics.png}
    \caption{$\mathrm{Re}=3{,}900$ --- RDE-PINN (function space): velocity diagnostics. The panels show the reference solution, ensemble mean prediction, absolute error, predictive standard deviation, and the error-to-uncertainty ratio for the velocity components $(U, V)$. The ensemble accurately reconstructs the velocity field, while the uncertainty estimates capture the spatial distribution of prediction error with reasonable calibration.}
    \label{fig:re3900_rde_fs_vel}
\end{figure}

\begin{figure}
    \centering
    \includegraphics[trim={0cm 7.9cm 0cm 0cm},clip,width=\linewidth]{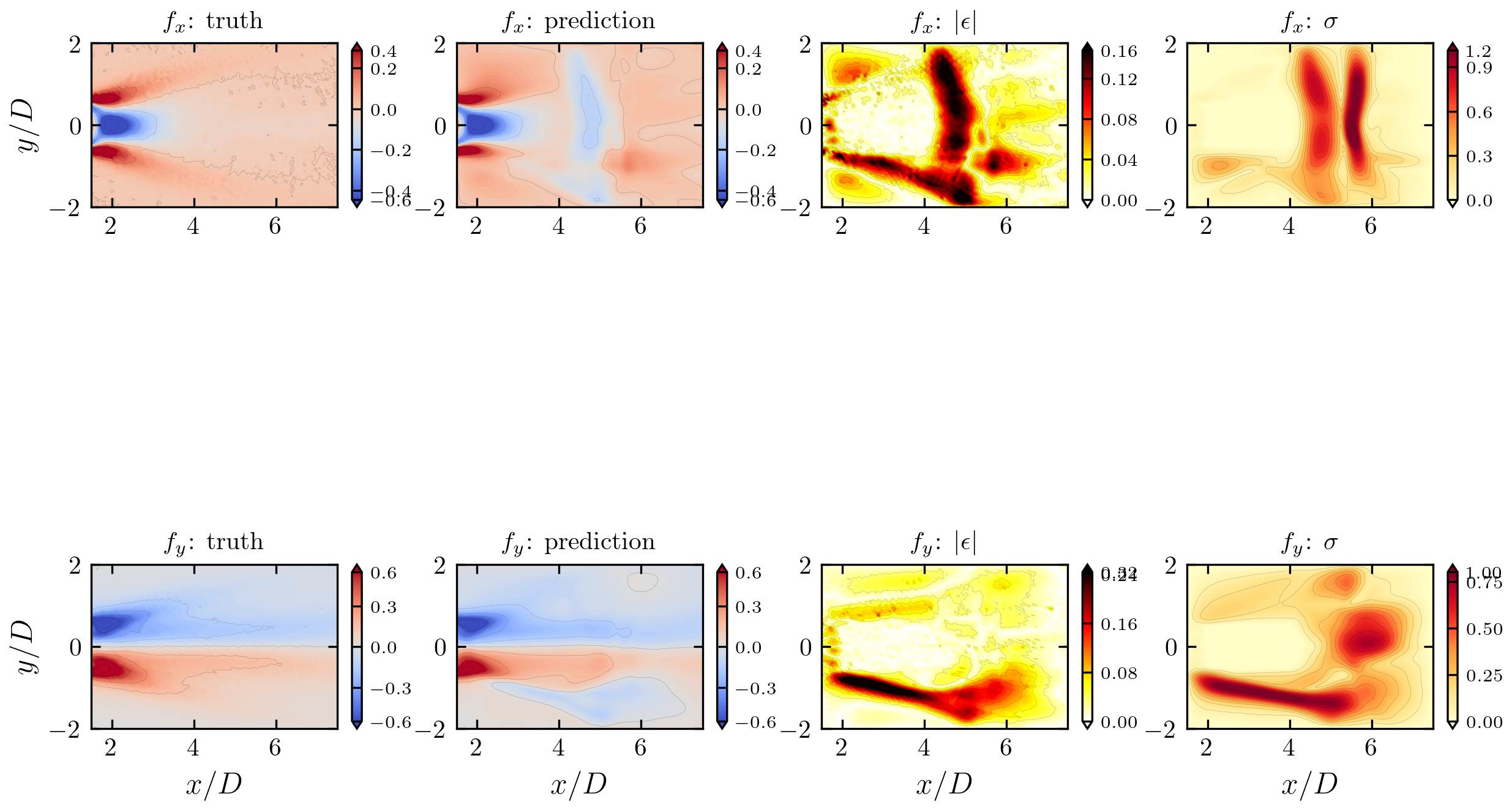}
    \includegraphics[trim={0cm 0cm 0cm 7cm},clip,width=\linewidth]{figs/RDE_Re_3900_FS/fig_reynolds_stress_diagnostics.png}
    \caption{$\mathrm{Re}=3{,}900$ --- RDE-PINN (function space): Reynolds stress divergence diagnostics. The panels present the reference fields, ensemble mean predictions, absolute error, predictive standard deviation, and the error-to-uncertainty ratio for the closure terms $(f_x, f_y)$. While the mean predictions capture the overall structure, the uncertainty estimates are less well calibrated, particularly in regions with large modeling error.}
    \label{fig:re3900_rde_fs_rs}
\end{figure}

\begin{figure}
    \centering
    \includegraphics[width=\linewidth]{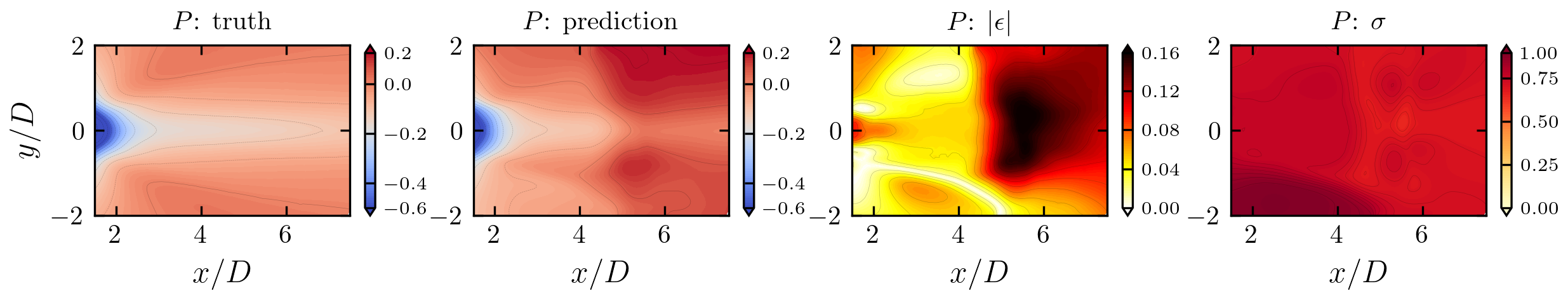}
    \caption{$\mathrm{Re}=3{,}900$ --- RDE-PINN (function space): inferred pressure field. No pressure data are used during training, and the pressure is recovered solely through the governing equations. The ensemble mean captures the large-scale structure of the pressure field; however, the associated uncertainty is relatively conservative, reflecting increased model variability in the absence of direct supervision.}
    \label{fig:re3900_rde_fs_pressure}
\end{figure}

The results in \autoref{fig:re3900_rde_fs_vel}--\autoref{fig:re3900_rde_fs_pressure} demonstrate that the function-space RDE-PINN provides accurate reconstructions of the mean velocity field and Reynolds forces, while also recovering the pressure field despite the absence of direct pressure supervision. The ensemble mean predictions show good agreement with the reference solutions across the domain for the velocity components $(U, V)$ (\autoref{fig:re3900_rde_fs_vel}) and the closure terms $(f_x, f_y)$ (\autoref{fig:re3900_rde_fs_rs}). In addition, the inferred pressure field (\autoref{fig:re3900_rde_fs_pressure}) captures the large-scale structure of the flow, highlighting the ability of the model to recover latent quantities purely from the governing equations. The ensemble standard deviation provides a spatially varying estimate of predictive uncertainty, which qualitatively aligns with regions of higher prediction error. This behavior is further supported by the error-to-uncertainty ratio fields, indicating that the ensemble captures the dominant error trends across both flow variables and closure terms. Moreover, the pairwise functional $L^2$ distances between ensemble members remain sufficiently large, confirming that the repulsive regularization promotes meaningful diversity in the predicted solutions rather than merely inducing separation in parameter space.

The quantitative performance of the function-space RDE-PINN is summarized in \autoref{tab:re3900_rde_fs}. For the primary flow variables, the model achieves low prediction errors, with RMSE values of $0.0138$ for $U$ and $0.0078$ for $V$, consistent with the accurate reconstructions observed in \autoref{fig:re3900_rde_fs_vel}. The associated uncertainty estimates are reasonably well calibrated, with coverage probabilities of $93.0\%$ and $96.3\%$ for the $\pm 2\sigma$ interval, respectively, and mean error-to-uncertainty ratios below unity, indicating slightly conservative uncertainty. For the pressure field, although no pressure data are used during training, the model recovers the global structure with moderate accuracy (RMSE $=0.1383$), as shown in \autoref{fig:re3900_rde_fs_pressure}. The corresponding uncertainty estimates are conservative, yielding full coverage ($100\%$) and a low error-to-uncertainty ratio.

In contrast, for the Reynolds stress divergence components $(f_x, f_y)$, the prediction errors are higher (RMSE $=0.1983$ and $0.3178$, respectively), as also evident in \autoref{fig:re3900_rde_fs_rs}. The associated uncertainty estimates are less well calibrated, with coverage probabilities ($75.2\%$ for $f_x$ and $70.4\%$ for $f_y$) falling below the nominal $95\%$ level and error-to-uncertainty ratios significantly larger than unity. This indicates underestimation of uncertainty in these quantities, reflecting the increased difficulty of modeling Reynolds stress dynamics. Overall, while the function-space RDE-PINN provides accurate predictions and reasonably calibrated uncertainty for the primary flow variables, its uncertainty estimates for the closure terms remain less reliable, highlighting an important limitation of the approach in capturing complex turbulent closure effects.

\begin{table}
  \centering
  \caption{Prediction accuracy and uncertainty calibration for Repulsive Deep Ensembles in function space at $\mathrm{Re}=3{,}900$.}
  \label{tab:re3900_rde_fs}
  \begin{tabular}{lcccccc}
    \toprule
    Variable & RMSE & $\bar{\sigma}_\mathrm{raw}$ & $\bar{\sigma}_\mathrm{cal}$ & $C_{2\sigma}^\mathrm{raw}$ & $C_{2\sigma}^\mathrm{cal}$ & $\langle|\epsilon|/\sigma_\mathrm{cal}\rangle$ \\
    \midrule
    $U$ & 0.0138 & 0.0593 & 0.0593 & 93.0\% & 93.0\% & 0.64 \\
    $V$ & 0.0078 & 0.0290 & 0.0290 & 96.3\% & 96.3\% & 0.48 \\
    $P$ & 0.1383 & 0.7480 & 0.7480 & 100.0\% & 100.0\% & 0.17 \\
    $f_x$ & 0.1983 & 0.1825 & 0.1825 & 75.2\% & 75.2\% & 11.77 \\
    $f_y$ & 0.3178 & 0.1977 & 0.1977 & 70.4\% & 70.4\% & 17.04 \\
    \bottomrule
  \end{tabular}
\end{table}

\paragraph{Comparison: Vanilla Deep Ensemble vs.\ RDE-PINN at 
Re\,=\,3{,}900}
In \autoref{tab:dns_dens} we report the prediction accuracy and uncertainty 
calibration of a deep ensemble PINN in which the repulsive term is omitted 
while all other hyperparameters remain identical to the RDE-PINN configuration 
(see \autoref{tab:re3900_rde_fs}). For the velocity fields, both methods 
achieve comparable RMSE ($1.49\times10^{-2}$ vs.\ $1.38\times10^{-2}$ for 
$U$; $6.78\times10^{-3}$ vs.\ $7.8\times10^{-3}$ for $V$), confirming that 
the repulsive term does not degrade predictive accuracy. However, uncertainty 
calibration differs dramatically: the vanilla ensemble yields 
$C_{2\sigma} = 65.0\%$ and $53.7\%$ with 
$\langle|\varepsilon|/\sigma\rangle = 2.15$ and $2.63$, whereas RDE-PINN 
recovers near-ideal coverage of $93.0\%$ and $96.3\%$ with 
$\langle|\varepsilon|/\sigma\rangle = 0.64$ and $0.48$, close to the 
theoretical value of $0.5$ for a well-calibrated Gaussian. For pressure, 
both methods exhibit large RMSE (${\sim}0.11$--$0.14$), consistent with $P$ 
being inferred purely from physics without direct supervision. The vanilla 
ensemble produces $\bar{\sigma} = 1.00\times10^{-1}$ with a spurious 
$C_{2\sigma} = 100\%$, while RDE-PINN yields $\bar{\sigma} = 0.748$, 
honestly reflecting the high epistemic uncertainty inherent in unsupervised 
pressure recovery, with $\langle|\varepsilon|/\sigma\rangle = 0.17$ 
confirming genuine conservative coverage. The most severe degradation occurs 
for the Reynolds stress closures $f_x$ and $f_y$, where the vanilla ensemble 
catastrophically collapses: $\bar{\sigma} \approx 1.38\times10^{-2}$ and 
$1.70\times10^{-2}$ are more than an order of magnitude below the respective 
RMSE values, yielding $C_{2\sigma} = 36.5\%$ and $1.8\%$ with 
$\langle|\varepsilon|/\sigma\rangle = 9.74$ and $20.43$. RDE-PINN 
substantially recovers diversity, with $\bar{\sigma} = 0.183$ and $0.198$ 
now matching the error scale, lifting $C_{2\sigma}$ to $75.2\%$ and $70.4\%$ 
and reducing $\langle|\varepsilon|/\sigma\rangle$ to $11.77$ and $17.04$. 
While still below the $95\%$ theoretical target, this represents a 
qualitative improvement in that the ensemble at least acknowledges uncertainty 
in the closure terms rather than collapsing to a single erroneous solution. 
Overall, without the repulsive kernel, random initialisation alone is 
insufficient to maintain ensemble diversity under the strong regularisation 
imposed by the shared RANS-PINN loss, and $C_{2\sigma}$ falls as low as 
$1.8\%$ against the $95\%$ target for $f_y$, making the vanilla uncertainty 
estimates physically meaningless for the quantities most critical to turbulence 
modelling.
\begin{table}
  \centering
  \caption{Prediction accuracy and uncertainty calibration for deep ensemble approach \cite{psaros2023uncertainty}}
  \label{tab:dns_dens}
  \begin{tabular}{lcccccc}
    \toprule
    Variable & RMSE & $\bar{\sigma}_\mathrm{raw}$ & $\bar{\sigma}_\mathrm{cal}$ & $C_{2\sigma}^\mathrm{raw}$ & $C_{2\sigma}^\mathrm{cal}$ & $\langle|\epsilon|/\sigma_\mathrm{cal}\rangle$ \\
    \midrule
    $U$ & 0.0149 & 0.0065 & 0.0065 & 65.0\% & 65.0\% & 2.15 \\
    $V$ & 0.0068 & 0.0030 & 0.0030 & 53.7\% & 53.7\% & 2.63 \\
    $P$ & 0.1093 & 0.1001 & 0.1001 & 100.0\% & 100.0\% & 1.08 \\
    $f_x$ & 0.1920 & 0.0138 & 0.0138 & 36.5\% & 36.5\% & 9.74 \\
    $f_y$ & 0.3182 & 0.0170 & 0.0170 & 1.8\% & 1.8\% & 20.43 \\
    \bottomrule
  \end{tabular}
\end{table}

\paragraph{Takeaway from DNS experiments}
The DNS-based experiments at $\mathrm{Re}=3{,}900$ reveal clear differences in the predictive accuracy and uncertainty quantification capabilities of BPINN and function-space RDE-PINN. As observed in \autoref{fig:re3900_bpinn_vel}--\autoref{fig:re3900_bpinn_pressure} and \autoref{fig:re3900_rde_fs_vel}--\autoref{fig:re3900_rde_fs_pressure}, both approaches are able to accurately reconstruct the mean velocity field and capture the large-scale flow features. However, the BPINN consistently provides better calibrated uncertainty estimates, with coverage close to the nominal $95\%$ level after recalibration across all variables, including the Reynolds stress divergence. In contrast, while the function-space RDE-PINN achieves comparable or even lower prediction errors for the primary flow variables, its uncertainty estimates exhibit mixed behavior: they are slightly conservative for velocity and pressure but significantly underconfident for the closure terms $(f_x, f_y)$, as evidenced by reduced coverage and elevated error-to-uncertainty ratios (\autoref{tab:re3900_rde_fs}).

A key advantage of BPINN lies in its fully Bayesian formulation, which enables consistent uncertainty propagation across all predicted quantities, including the pressure field that is inferred purely from the governing equations. The resulting uncertainty estimates remain well aligned with the true error even in regions without direct supervision. In contrast, the RDE-PINN relies on ensemble diversity induced through function-space repulsion, which effectively captures dominant error trends for primary variables but does not guaranty reliable calibration for more complex quantities, such as Reynolds stress divergence. The importance of the function-space repulsive kernel becomes further evident when RDE-PINN is compared against a vanilla deep ensemble PINN trained under identical hyperparameters but without the repulsion term (see \autoref{tab:dns_dens}). Without repulsion, random initialisation alone  is insufficient to maintain ensemble diversity under the strong regularisation imposed by the shared RANS-PINN loss, causing all members to collapse onto a single solution. This collapse is most catastrophic for the Reynolds stress closures, where $C_{2\sigma}$ falls to $1.8\%$ for $f_y$ and $\langle|\varepsilon|/\sigma\rangle$ reaches $20.43$, against $70.4\%$ and $17.04$ for RDE-PINN and confirming that the repulsive kernel is the decisive ingredient for recovering meaningful ensemble diversity, even if reliable calibration for closure terms remains an open challenge shared by all ensemble-based PINN approaches. 

Overall, these results suggest that BPINN provides more robust and trustworthy uncertainty quantification for turbulent flow inference, particularly for ill-posed quantities and closure terms. The function-space RDE-PINN, on the other hand, offers a computationally efficient alternative with strong predictive accuracy and reasonable uncertainty estimates for primary variables, but its limitations in calibrating uncertainty for closure dynamics highlight an important direction for further improvement.

% ------------------------------------------------------------------
\subsubsection{$\mathrm{Re} = 10{,}000$: Experimental Data}
\label{sec:re10000}
\paragraph{PIV Data generation and data used for UQ analysis}
The PIV experiment was performed using the robotic tow tank facility at the Massachusetts Institute of Technology.
The facility has a length of 10 m and a cross-section of 1 m $\times$ 1 m~\cite{fan2019robotic}. To generate the PIV data, a $d_\mathrm{c}=50.8$ mm diameter circular cylinder was towed by a gantry robot running on top of the towing tank. To visualize the flow field, the water inside the tow tank was seeded with 5 $\mu m$ polyamide particles, which exhibited neutral buoyancy throughout the experiment. A horizontal light sheet, perpendicular to the cylinder's main axis, was generated by a laser (Optolutions LD-PS/40) mounted on the gantry robot but positioned outside the towing tank. The camera (Optronis Cyclone 2000) and lens (Zeiss Dimension 2/35) were mounted behind the cylinder, inside a waterproof housing. All components, the cylinder, the camera, and the laser, were moving together at the towing velocity $U_{\mathrm{tow}}= 0.188$ m/s. To compensate for optical distortions and to obtain the scaling factor $s_\mathrm{p}=5.744$ pixels/mm, a camera calibration was performed.
To obtain well-converged statistics, five experimental runs were performed, yielding approximately 30,000 individual flow fields. The images were recorded as a time series at a frame rate of 250 Hz.
The PIV processing was performed using the DAVIS 11 software with an interrogation window size of 16 pixel$\times$16 pixel and 75$\%$ overlap. Outliers were removed using the universal outlier detection method~\cite{westerweel2005universal}. As a result, flow fields on a grid of 301$\times$220 vectors were obtained and subsequently non-dimensionalized using the $d_\mathrm{c}$ and $U_{\mathrm{tow}}$ as reference.

While the random uncertainty of the mean fields is low due to the averaging over 30,000 snapshots, the mean fields are still impacted by measurement errors. These errors can never be fully eliminated, even for the most carefully performed experiments. In the context of this manuscript, a few error sources are listed below:
\begin{itemize}
    \item Missalignment of the light sheet plane and the calibration plane 
    \item Misalignment between the cylinder and the light sheet plane
    \item Low-pass filter effect of PIV averaging over windows
\end{itemize}

\begin{figure}[htbp]
    \centering
    \begin{minipage}{0.48\linewidth}
        \centering
        \includegraphics[trim={1cm 0cm 3cm 1cm}, clip, width=\linewidth]{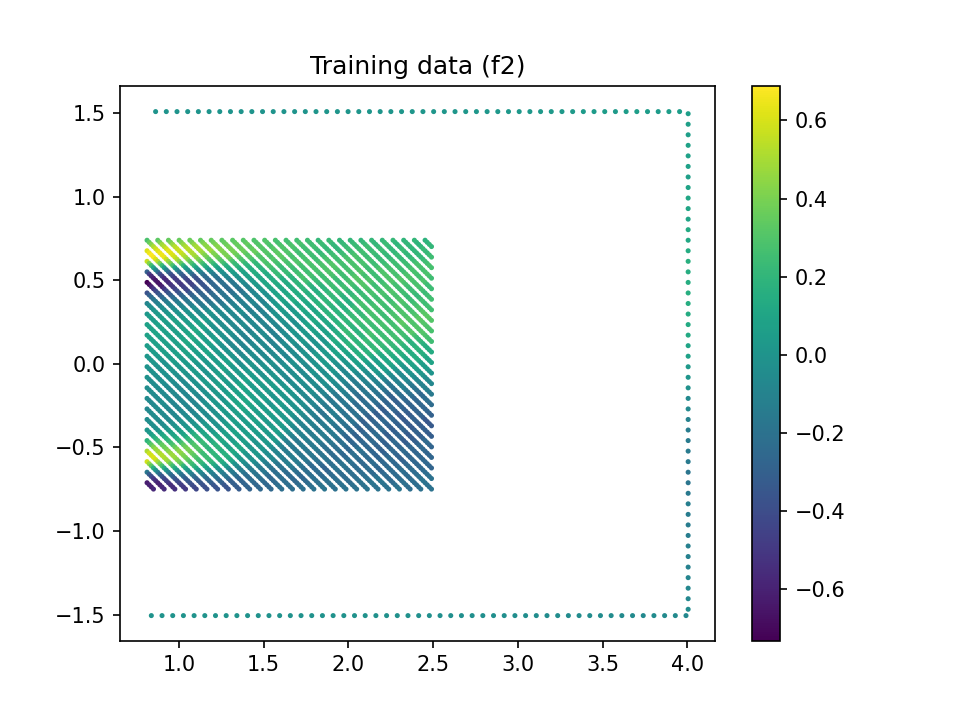}
    \end{minipage}
    \hfill
    \begin{minipage}{0.48\linewidth}
        \caption{$\mathrm{Re}=10{,}000$: Computational domain and data used for training and inference. The plotted points indicate the sparse measurement locations used for uncertainty quantification (UQ). Observations are available only for the velocity components $(U, V)$ and the closure terms $(f_x, f_y)$, while no pressure data are used during training, reflecting the practical limitation that pressure measurements are typically unavailable in experiments. Color shows the distribution $f_x$.}
        \label{fig:re10k_domain}
    \end{minipage}
\end{figure}

We consider the turbulent flow past a circular cylinder at $\mathrm{Re}=10{,}000$, where the objective is to infer the flow field and associated closure quantities from sparse and noisy observations. The computational setup is illustrated in \autoref{fig:re10k_domain}. A limited number of measurement locations are distributed across the domain to emulate a data-scarce experimental scenario. Observations are available only for the mean velocity components $(U, V)$ and the closure terms $(f_x, f_y)$, while no pressure data are used during training, reflecting the practical difficulty of obtaining pressure measurements in experiments. Gaussian noise is added to all observed quantities to account for measurement uncertainty. In addition, a set of collocation points is used to enforce the governing equations, while a dense set of test points is employed to evaluate predictive accuracy and uncertainty quantification across the domain. This setup provides a challenging benchmark for assessing the robustness and calibration of the proposed UQ methods under realistic, data-limited conditions.

\paragraph{Bayesian PINN.}
We consider uncertainty quantification for turbulent flow past a circular cylinder at $\mathrm{Re}=10{,}000$ using a BPINN trained on particle image velocimetry (PIV)-type data. The surrogate model is a fully connected neural network with architecture $[2, 64, 64, 64, 64, 64, 5]$, mapping spatial coordinates $(x,y)$ to the outputs $(U, V, P, f_x, f_y)$. Training data consist of sparse and noisy measurements of the velocity components $(U, V)$ and the closure terms $(f_x, f_y)$, while no pressure data are used during training; instead, the pressure field is inferred solely through the governing equations. The Reynolds-averaged Navier--Stokes (RANS) equations are enforced via a physics-informed residual formulation, with collocation points distributed across the domain. 
The training procedure employs a three-stage maximum a posteriori (MAP) pre-training strategy, consisting of a data-only phase, a gradual incorporation of physics constraints through a cosine ramp, and a final LBFGS-based refinement. Subsequently, full Bayesian inference is performed using the NUTS, with subsampling of collocation points to maintain computational tractability. A tempered multi-component likelihood is used to balance the contributions of velocity data, Reynolds stress data, and PDE residuals, improving both sampling efficiency and posterior calibration. To address the well-known underestimation of uncertainty in Bayesian neural networks, a post-hoc recalibration procedure is applied to the posterior standard deviation, ensuring that predictive intervals achieve the desired coverage.

\begin{table}[htbp]
  \centering
  \caption{Prediction accuracy and uncertainty calibration for the Bayesian PINN at $\mathrm{Re}=10{,}000$.}
  \label{tab:re10k_bpinn}
  \begin{tabular}{lcccccc}
    \toprule
    Variable & RMSE & $\bar{\sigma}_\mathrm{raw}$ & $\bar{\sigma}_\mathrm{cal}$ & $C_{2\sigma}^\mathrm{raw}$ & $C_{2\sigma}^\mathrm{cal}$ & $\langle|\epsilon|/\sigma_\mathrm{cal}\rangle$ \\
    \midrule
    $U$ & 0.0277 & 0.0085 & 0.0304 & 57.6\% & 95.0\% & 0.66 \\
    $V$ & 0.0108 & 0.0060 & 0.0086 & 76.9\% & 95.0\% & 0.90 \\
    $f_x$ & 0.0398 & 0.0146 & 0.0587 & 60.0\% & 95.0\% & 0.60 \\
    $f_y$ & 0.0550 & 0.0140 & 0.0597 & 44.4\% & 95.0\% & 0.69 \\
    \bottomrule
  \end{tabular}
\end{table}

\begin{figure}
    \centering
    \includegraphics[height=5cm]{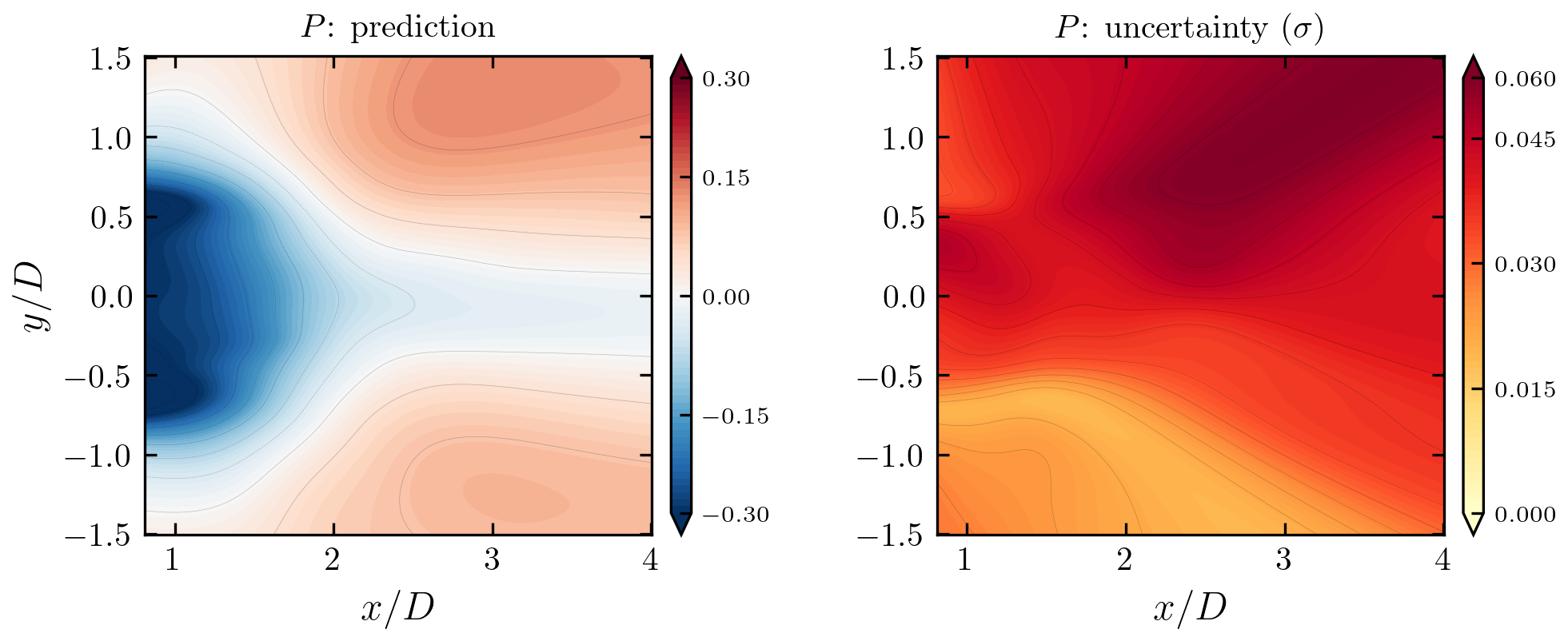}
    \caption{$\mathrm{Re}=10{,}000$ --- Bayesian PINN: inferred mean pressure field ($P$). No pressure data are used during training, and the pressure is recovered solely through the governing equations. The posterior mean captures the large-scale structure of the pressure field, while the predictive uncertainty provides a measure of confidence in the absence of direct observations.}
    \label{fig:re10k_pressure}
\end{figure}

The results for the BPINNs method at $Re=10{,}000$ (see \autoref{fig:re10k_vel}--\autoref{fig:re10k_pressure} and \autoref{tab:re10k_bpinn}) demonstrate accurate reconstruction of the mean flow quantities together with reliable uncertainty quantification after recalibration. The posterior mean predictions closely match the reference velocity field, with low errors ($\mathrm{RMSE}_u = 0.0277$, $\mathrm{RMSE}_v = 0.0108$), while the Reynolds stress divergence components $(f_x, f_y)$ are recovered with moderately higher error ($0.0398$ and $0.0550$, respectively), reflecting the increased difficulty of inferring closure terms from indirect constraints. Despite the absence of pressure data, the BPINN successfully reconstructs the pressure field in a physically consistent manner (\autoref{fig:re10k_pressure}). The raw posterior uncertainty obtained from NUTS is systematically underconfident, with coverage ranging from $44.4\%$ to $76.9\%$, consistent with known limitations of Bayesian neural networks in high-dimensional settings. Following post-hoc recalibration, the predictive uncertainty achieves the nominal $95\%$ coverage across all variables, with recalibrated standard deviations becoming commensurate with the observed errors. The resulting error-to-uncertainty ratios indicate near-ideal calibration for the transverse velocity ($0.90$) and slightly conservative estimates for the remaining variables ($0.60$--$0.69$), which is desirable in practical applications. Spatially, the uncertainty fields correctly identify regions of elevated error, particularly in the wake and shear-layer regions (\autoref{fig:re10k_vel}), where turbulent effects are strongest. The Reynolds stress divergence exhibits larger uncertainty and lower raw coverage due to its indirect dependence on the governing equations, but its spatial uncertainty structure remains physically meaningful, with peaks aligned with regions of high shear and turbulence production. Overall, these results highlight the robustness of the BPINN framework for uncertainty quantification in realistic, data-limited turbulent flow scenarios, particularly when only partial observations such as PIV measurements are available.

\begin{figure}
    \centering
    \includegraphics[trim={0cm 7.9cm 0cm 0cm},clip,width=\linewidth]{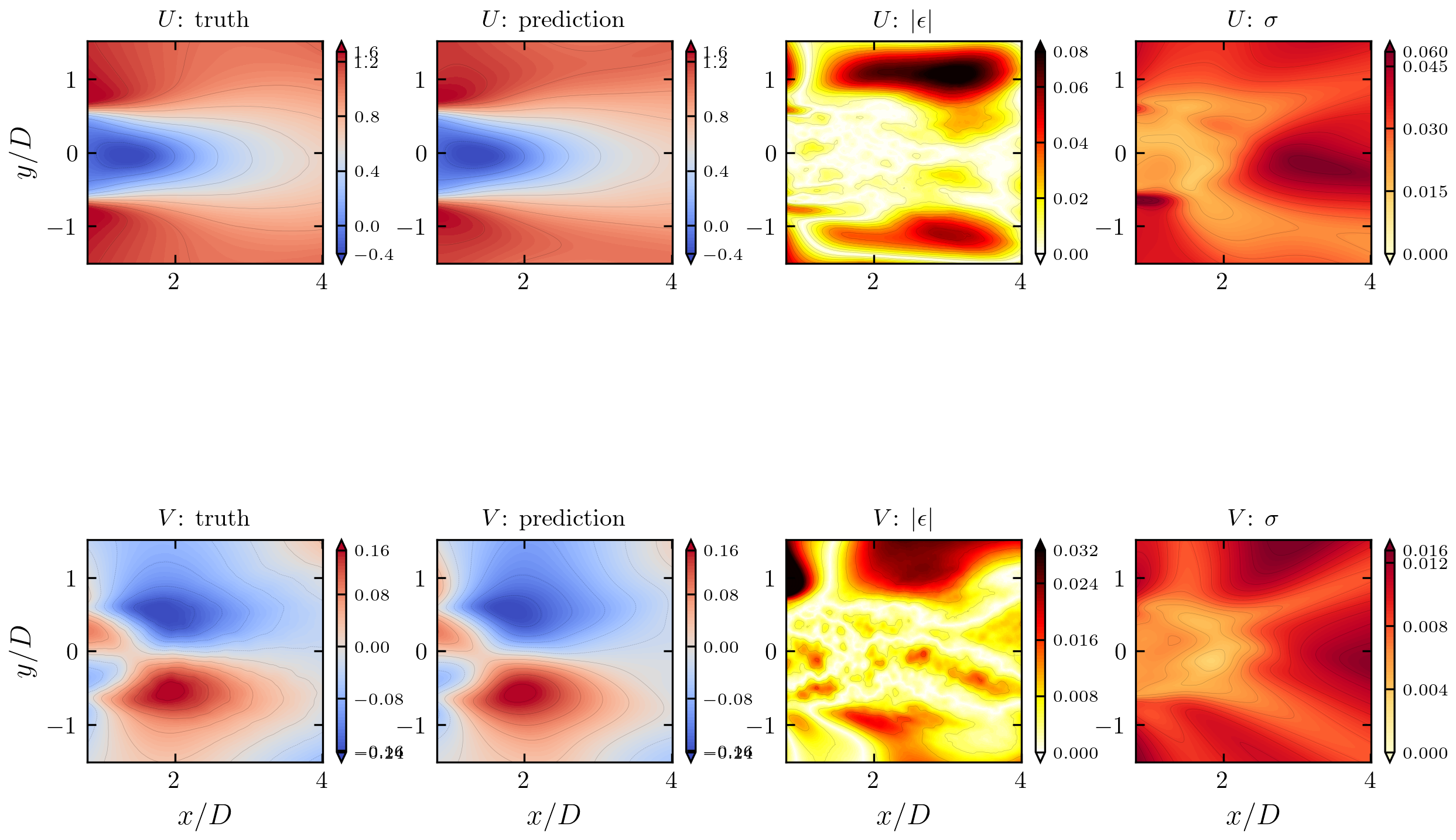}
    \includegraphics[trim={0cm 0cm 0cm 7cm},clip,width=\linewidth]{figs/Re_10000/fig_velocity_diagnostics.png}
    \caption{$\mathrm{Re}=10{,}000$ --- Bayesian PINN velocity diagnostics. The panels show the reference solution, posterior mean prediction, absolute error, and predictive standard deviation, for the velocity components $(U, V)$. The results demonstrate accurate reconstruction of the velocity field and good alignment between predicted uncertainty and pointwise error across the domain after recalibration.}
    \label{fig:re10k_vel}
\end{figure}

\begin{figure}
    \centering
    \includegraphics[trim={0cm 7.9cm 0cm 0cm},clip,width=\linewidth]{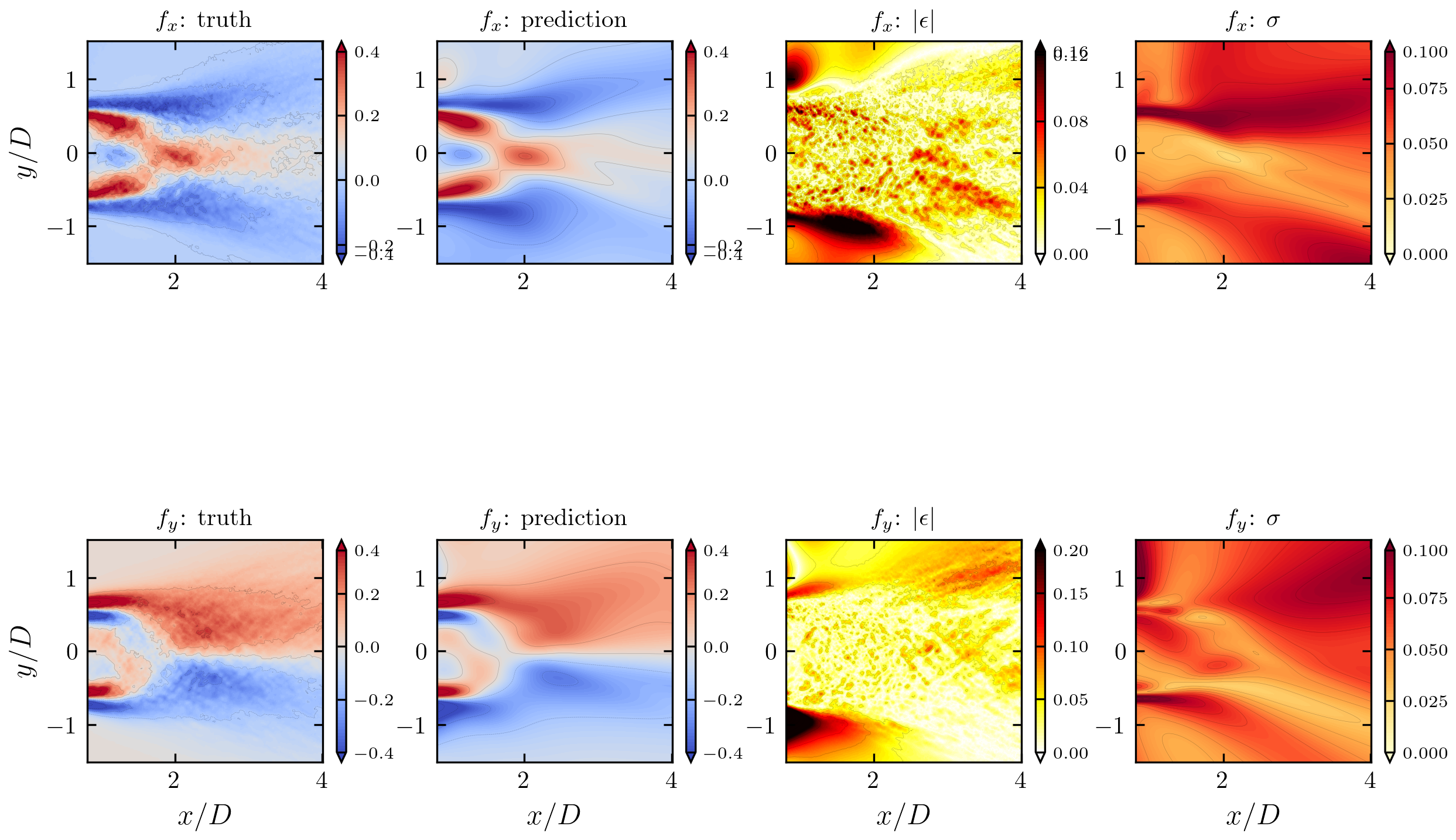}
    \includegraphics[trim={0cm 0cm 0cm 7cm},clip,width=\linewidth]{figs/Re_10000/fig_reynolds_stress_diagnostics.png}
    \caption{$\mathrm{Re}=10{,}000$ --- Bayesian PINN Reynolds stress diagnostics. The panels present the reference fields, posterior mean predictions, absolute error, and predictive standard deviation for the closure terms $(f_x, f_y)$. While the mean predictions capture the overall structure, the uncertainty estimates reflect increased modeling difficulty in the closure terms.}
    \label{fig:re10k_rs}
\end{figure}

\paragraph{Repulsive Deep Ensembles in Function Space.}
For this case, the network architecture, loss formulation, and training procedure remain identical to the $\mathrm{Re} = 3{,}900$ configuration, with the sole modification that the kinematic viscosity is set to $\mathrm{Re} = 10^{4}$.

\begin{figure}
    \centering
    \includegraphics[trim={0cm 7.9cm 0cm 0cm},clip,width=\linewidth]{figs/Re_10000/fig_velocity_diagnostics.png}
    \includegraphics[trim={0cm 0cm 0cm 7cm},clip,width=\linewidth]{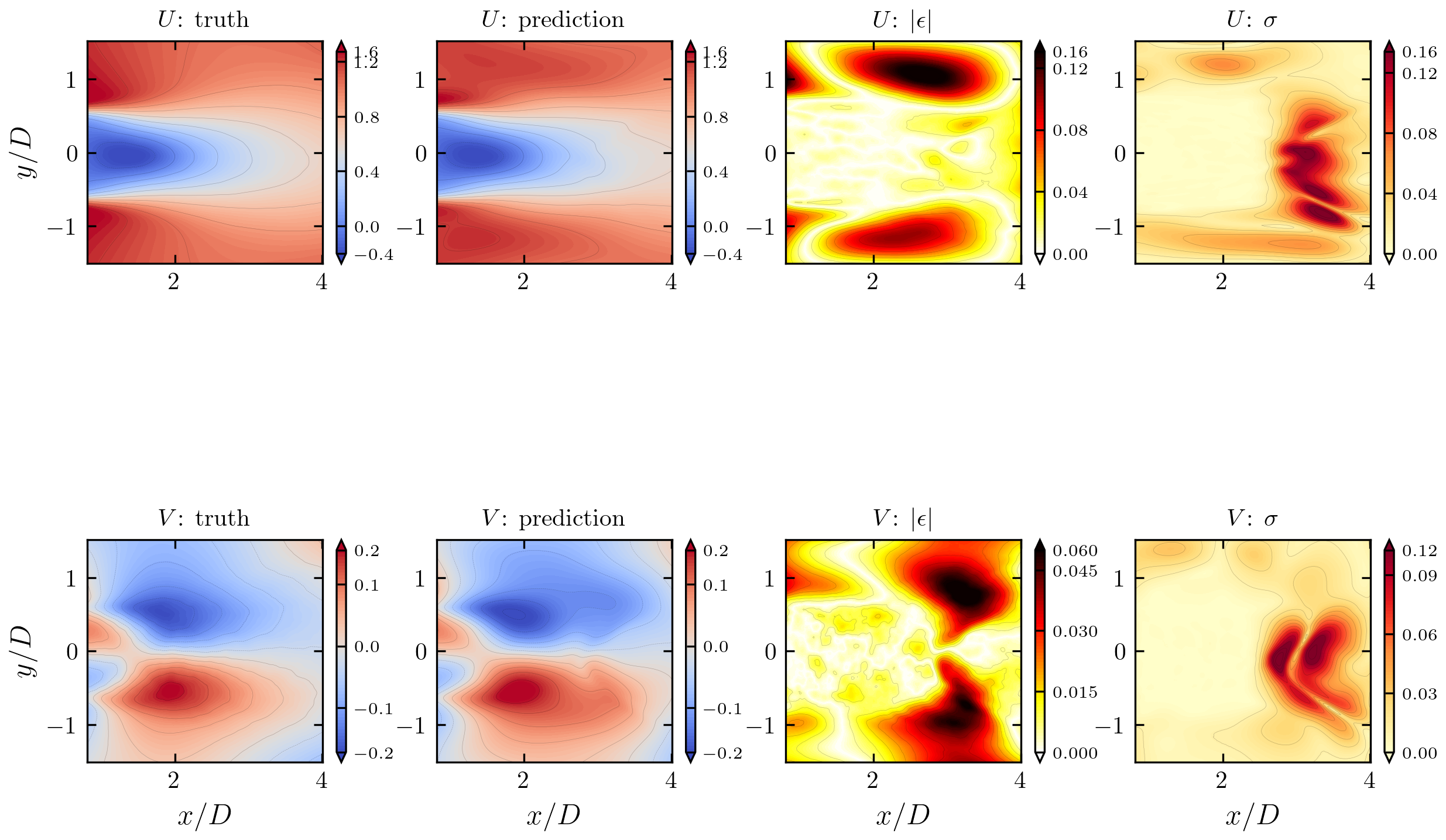}
    \caption{$\mathrm{Re}=10{,}000$ ---  RDE-PINN  velocity diagnostics. The panels show the reference solution, posterior mean prediction, absolute error, and predictive standard deviation, for the velocity components $(U, V)$. The results demonstrate accurate reconstruction of the velocity field and good alignment between predicted uncertainty and pointwise error across the domain after recalibration.}
    \label{fig:re10k_vel}
\end{figure}

\begin{figure}
    \centering
    \includegraphics[trim={0cm 7.9cm 0cm 0cm},clip,width=\linewidth]{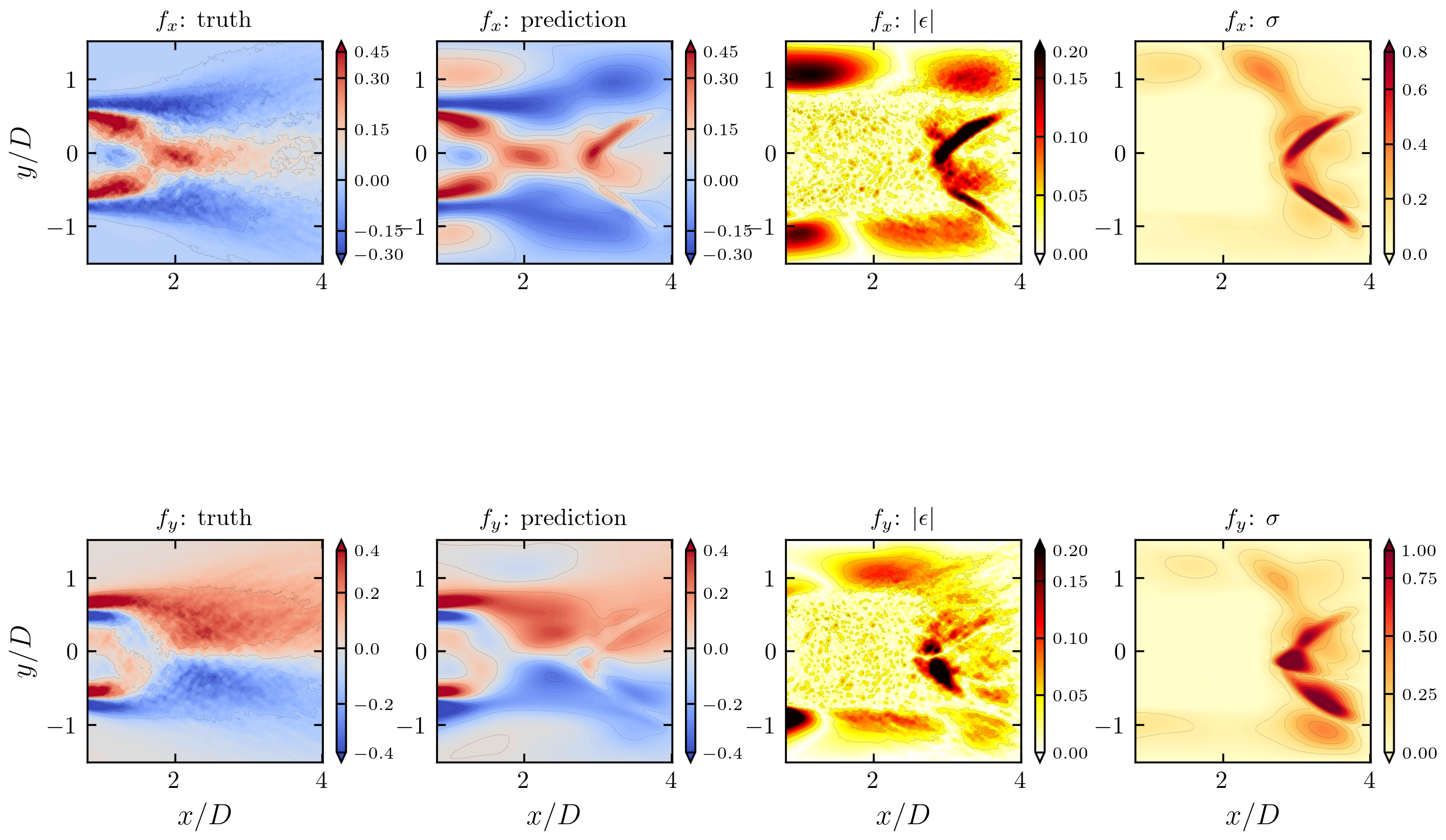}
    \includegraphics[trim={0cm 0cm 0cm 7cm},clip,width=\linewidth]{figs/Re_10K_RDE/fig_reynolds_stress_diagnostics.png}
    \caption{$\mathrm{Re}=10{,}000$ --- RDE-PINN Reynolds stress diagnostics. The panels present the reference fields, posterior mean predictions, absolute error, and predictive standard deviation for the closure terms $(f_x, f_y)$. While the mean predictions capture the overall structure, the uncertainty estimates reflect increased modeling difficulty in the closure terms.}
    \label{fig:re10k_rs}
\end{figure}

\begin{table}
  \centering
  \caption{Prediction accuracy and uncertainty calibration for the RDE-PINN at $\mathrm{Re}=10{,}000$.}
  \label{tab:results}
  \begin{tabular}{lcccccc}
    \toprule
    Variable & RMSE & $\bar{\sigma}_\mathrm{raw}$ & $\bar{\sigma}_\mathrm{cal}$ & $C_{2\sigma}^\mathrm{raw}$ & $C_{2\sigma}^\mathrm{cal}$ & $\langle|\epsilon|/\sigma_\mathrm{cal}\rangle$ \\
    \midrule
    $U$ & 0.0462 & 0.0290 & 0.0290 & 62.3\% & 62.3\% & 2.07 \\
    $V$ & 0.0205 & 0.0189 & 0.0189 & 65.1\% & 65.1\% & 1.99 \\
    $f_x$ & 0.0590 & 0.0889 & 0.0889 & 78.4\% & 78.4\% & 1.56 \\
    $f_y$ & 0.0471 & 0.1128 & 0.1128 & 81.5\% & 81.5\% & 1.33 \\
    \bottomrule
  \end{tabular}
\end{table}

% ====================================================================
\section{Summary}
This work presents a systematic comparison of three uncertainty quantification (UQ) methods for physics-informed turbulence modeling \cite{ patel2024turbulence, eivazi2022physics, zhang2025turbulence}: (i) Bayesian PINNs \cite{yang2021b} with tempered No-U-Turn sampling \cite{hoffman2014no} and post-hoc recalibration \cite{kuleshov2018accurate}, (ii) MC Dropout PINNs \cite{zhang2019quantifying}, and (iii) Repulsive Deep Ensemble PINNs with both function-space and parameter-space repulsion. Through a progression of test cases---from the Van der Pol oscillator to turbulent flow past a cylinder at $\mathrm{Re}=3{,}900$ (DNS) and $\mathrm{Re}=10{,}000$ (experimental PIV)---we demonstrate that the Bayesian PINN provides the most reliable and well-calibrated uncertainty estimates across all inferred quantities, including velocity, pressure (recovered without direct observations), and Reynolds stress divergence, achieving nominal $95\%$ coverage after recalibration. The function-space Repulsive Deep Ensemble offers a computationally efficient alternative with strong predictive accuracy for primary flow variables; however, its uncertainty calibration degrades for the closure terms. In contrast, parameter-space repulsion and MC Dropout are found to be less effective and are not recommended for ill-posed RANS inverse problems. 

Overall, these findings provide practical guidance for selecting UQ methods in data-driven turbulence modeling: Bayesian PINNs when calibration fidelity is critical, function-space repulsive ensembles when computational efficiency is required.
\label{sec:conclusions}

\section{Acknowledgments}
This research was supported by the Defense Advanced Research Projects Agency (DARPA) through the Automated Prediction Aided by Quantized Simulators (APAQuS) program under Grant No.~HR00112490526. KS would like to acknowledge \textbf{Prabhjyot Singh Saluja} (CCV, Brown) for providing the hardware expertise and support for the AI testbed used to benchmark the applications presented in the paper.

\appendix
\section{Deep ensemble approach for Van der Pol oscillator} \label{app:vpol_dens}
\begin{figure}
    \centering
    \includegraphics[width=\linewidth]{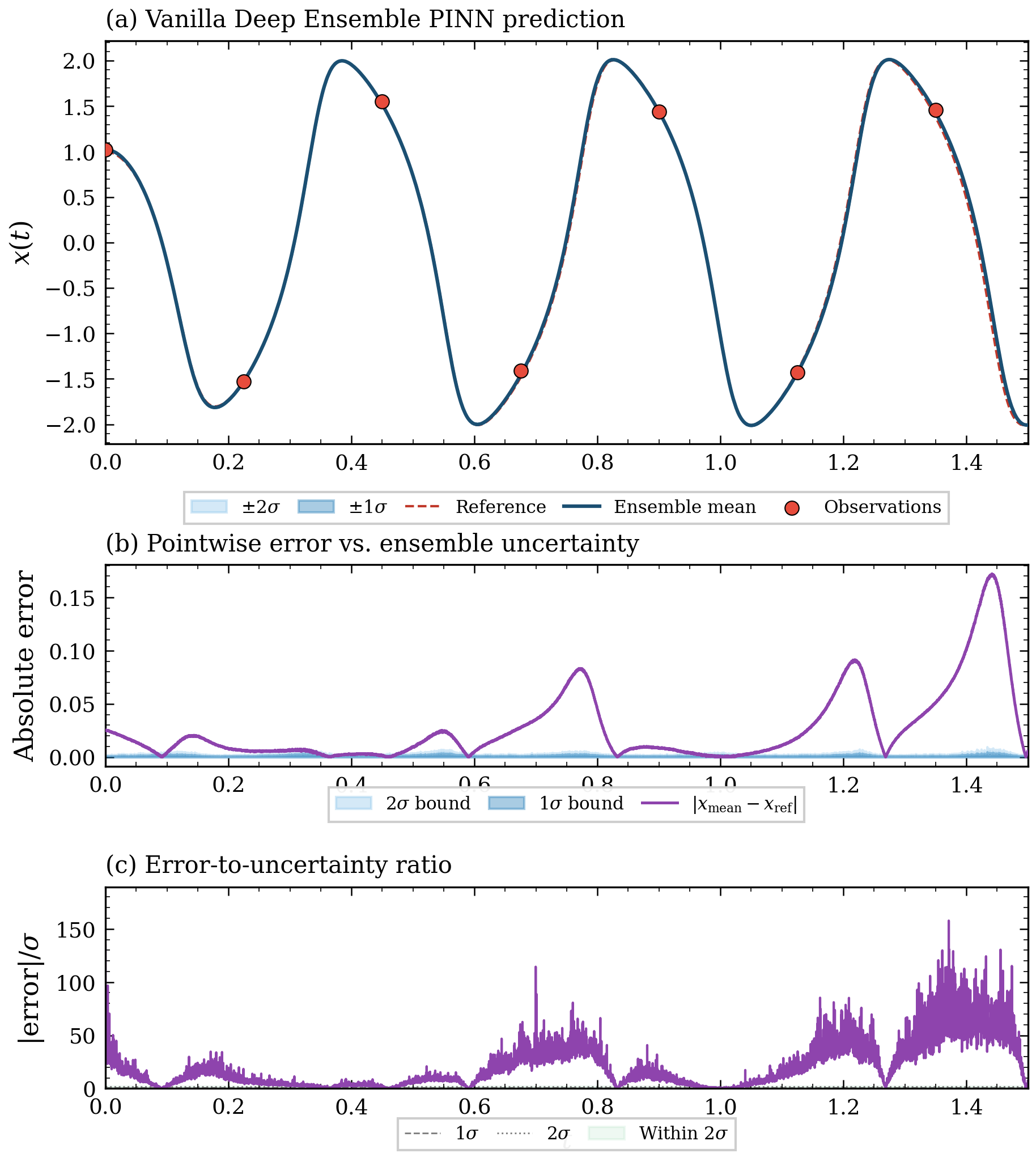}
    \caption{Uncertainty quantification of Van der Pol oscillator using Deep Ensemble approach \cite{psaros2023uncertainty} with out any repulsion terms.}
    \label{fig:dens_vpol}
\end{figure}

In \autoref{fig:dens_vpol}, we present the results obtained using the Deep Ensemble approach for uncertainty quantification of the Van der Pol oscillator. In \autoref{fig:dens_vpol}(a) the ensemble mean tracks the reference solution well,
but the $\pm 1\sigma$ and $\pm 2\sigma$ bands are completely invisible and all $M=10$ members have collapsed onto a single solution, producing accurate predictions but \textbf{no meaningful uncertainty}. In \autoref{fig:dens_vpol}(b) the absolute error
$|\bar{x}_{\mathrm{mean}} - x_{\mathrm{ref}}|$ grows significantly for
$t > 0.6$, peaking near $0.17$ at $t \approx 1.45$. The uncertainty bounds
remain \textbf{flat at zero} throughout. The ensemble is most wrong precisely
where it is most confident and shows a direct failure of calibration. In \autoref{fig:dens_vpol}(c)The ratio $|\mathrm{error}|/\sigma$ exceeds $100$ across
much of the domain, with spikes above $150$ near $t \approx 1.45$. A
well-calibrated ensemble requires $|\mathrm{error}|/\sigma \leq 1$ for
${\sim}68\%$ of points; here this condition is violated \textbf{everywhere}. Therefore without a repulsion mechanism, gradient descent further homogenized by the shared PDE loss drives all members into the
\textbf{same loss basin}, collapsing ensemble diversity. Coverage within $\pm 2\sigma$ is effectively $0\%$ against a theoretical target of ${\sim}95\%$, confirming that the Vanilla Deep Ensemble degenerates into an overconfident point estimator. This motivates the RDE-PINN repulsive kernel, which explicitly penalizes functional similarity to restore calibrated
uncertainty.

% ====================================================================
% Bibliography
% ====================================================================
\newpage
\bibliographystyle{elsarticle-num}
\bibliography{reference}  % Uncomment when you have a a

\end{document}